%% file: main.tex
\documentclass[10pt,twocolumn,letterpaper]{article}

\usepackage{cvpr}              
\usepackage{float}
\usepackage{graphicx}

\input{preamble}

\definecolor{cvprblue}{rgb}{0.21,0.49,0.74}
\usepackage[pagebackref,breaklinks,colorlinks,allcolors=cvprblue]{hyperref}

\def\confName{CVPR}
\def\confYear{2027}

\title{Points2Medial: Feed-Forward Medial Fields from Point Clouds}

\author{%
\begin{tabular}{c@{\hspace{4em}}c}
Martín Sinnona$^{1,3}$ & Emmanuel Iarussi$^{1,2}$ \\
{\tt\small martin.sinnona@utdt.edu} & {\tt\small emmanuel.iarussi@utdt.edu}
\end{tabular}\\[1.5em]
$^{1}$Laboratorio de Inteligencia Artificial, Universidad Torcuato Di Tella, Buenos Aires, Argentina. \\  
$^{2}$Consejo Nacional de Investigaciones Científicas y Técnicas, Buenos Aires, Argentina. \\ 
$^{3}$Facultad de Ciencias Exactas y Naturales, Universidad de Buenos Aires, Argentina.
}

\begin{document}
\maketitle

\input{sections/abstract}

\input{sections/intro}
\input{sections/related}
\input{sections/background}
\input{sections/method}

\input{sections/results}

\input{sections/limitations}
\input{sections/conclusion}
\input{sections/AI_statement}
\input{sections/acknowledgements}

\small
\bibliographystyle{ieeenat_fullname}
\bibliography{main}

\input{sections/supplementary}

\end{document}

%% file: preamble.tex
\usepackage{multirow}



%% file: sections/abstract.tex
\begin{abstract}
The medial axis transform describes a solid through its internal
skeleton and associated radii, providing a compact representation
for shape analysis, reconstruction, and geometric modeling.
Neural medial fields make this information accessible through
continuous functions, but existing approaches require fitting a
separate network to each shape. We introduce Points2Medial, a
feed-forward model that predicts continuous signed distance and
medial fields from sparse oriented point clouds, generalizing
to unseen shapes without test-time network fitting. Learning
this mapping across shapes requires reasoning beyond local
surface geometry: similar patches can belong to shapes with
very different thicknesses. Our key idea is to associate interior
queries with estimated surface points, or anchors. At each
anchor, the model combines local geometry with global shape
context to predict the radius of a maximal inscribed sphere
tangent to the surface at that point. Training uses
geometric constraints without requiring ground-truth medial
axes or radii. The predicted fields
jointly define an implicit medial axis from which a medial mesh
can be extracted. Points2Medial achieves lower mean medial-axis
Chamfer and Hausdorff distances on ABC than those reported for
Q-MDF, and comparable accuracy on Thingi10K despite being trained
only on ABC. As a by-product, our SDF branch reduces mean
signed-distance error deep inside ABC shapes by a factor of
6--8 compared with Points2Surf.
\end{abstract}

%% file: sections/intro.tex
\section{Introduction}
\label{sec:intro}

\begin{figure}[t]
    \centering
    \includegraphics[width=\columnwidth]{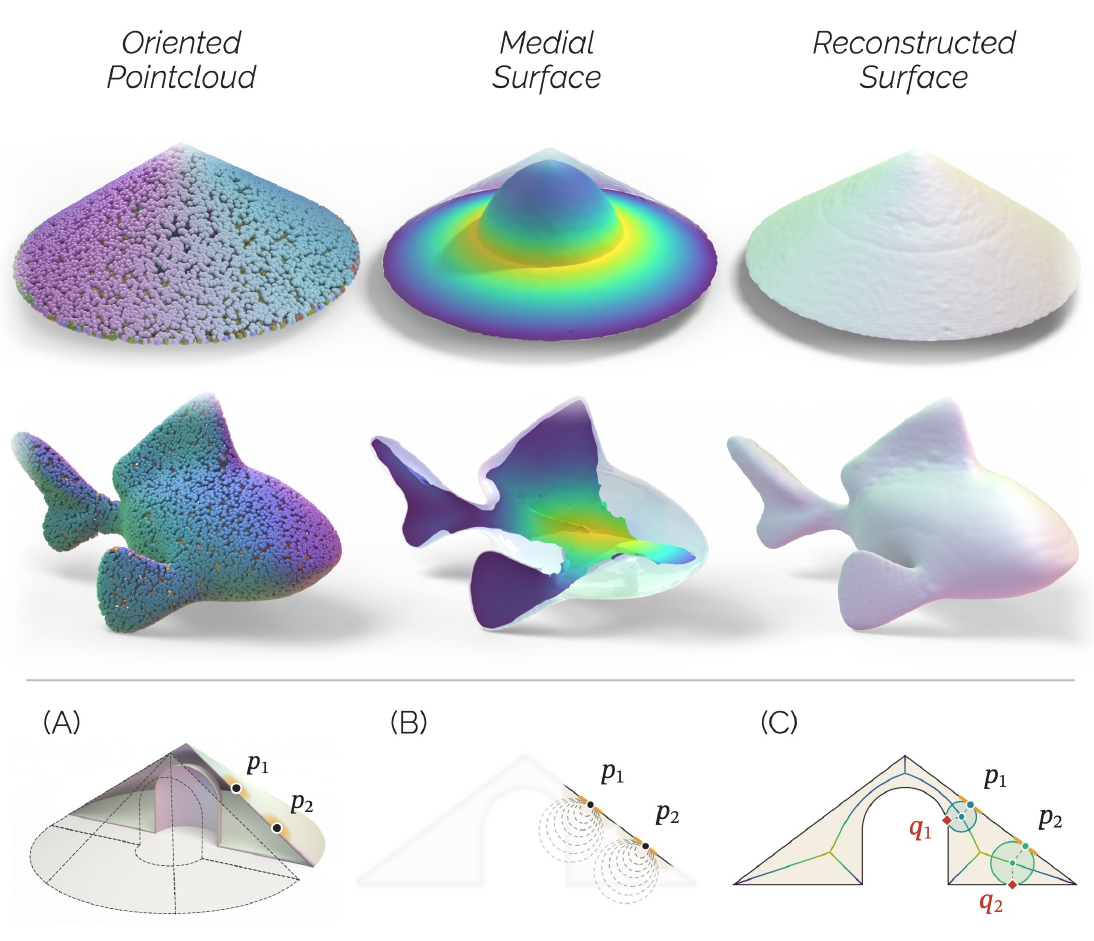}
    \caption{\textbf{Points2Medial.}
\emph{Top:} From an oriented point cloud, we reconstruct a medial surface and its associated radii, providing a compact representation from which the shape's surface can be recovered.
\qquad
\emph{Bottom:}
(\textbf{A}) Cutaway view of a shape, with two marked surface points $p_1$ and $p_2$.
(\textbf{B}) Local geometry alone cannot uniquely determine the radius of the maximal inscribed sphere tangent at either point.
(\textbf{C}) The medial radius depends on other boundary
regions: the sphere tangent at $p_1$ is constrained by the opposite wall at $q_1$, while the sphere at $p_2$ is constrained by the base at $q_2$. Points2Medial combines local geometry with global shape context to predict radii at surface points, yielding a continuous medial field used to extract a medial mesh.}
    \label{fig:teaser}
\end{figure}

The medial axis describes a shape from the inside. Its points are the centers of maximal inscribed spheres, whose radii measure local thickness. Together, these centers and radii form the medial axis transform, a complete representation of the solid from which its surface can be recovered (Fig.~\ref{fig:teaser}, top). This representation makes relationships between different parts of the surface explicit, supporting applications in shape analysis, geometric modeling,
and shape generation. Recovering it from a point cloud, however, remains difficult: the surface is known only through discrete samples, and small changes in its geometry can produce large changes in the medial axis.

Recent neural methods address this problem by representing medial geometry through continuous fields. They jointly estimate a signed distance field, which describes the surface, and a medial field, which encodes the radii of associated medial spheres. These fields provide an implicit representation from which a medial mesh can be extracted. However, they require fitting a separate network for each shape. Processing a new observation therefore means repeating the optimization. Our goal is to learn this mapping across shapes, so that a new point cloud can be processed without fitting another network.

Surface reconstruction offers a useful starting point. Shapes with different overall structures often share similar local surface patches, allowing a network to learn geometric features that transfer between objects. For medial prediction, however, similar patches can correspond to different radii. Consider two plates with different thicknesses. A small patch on one side may look identical in both, yet the corresponding medial radii differ because the opposite sides lie at different distances. Predicting a continuous medial field therefore requires relating different regions of the boundary. The predictions must also remain consistent throughout the interior: points along a medial spoke—the segment joining a boundary point to its associated medial center—share the same radius.

We present Points2Medial, a feed-forward network that predicts
signed distance and medial fields from sparse oriented point
clouds. Separate SDF and medial branches combine local surface
geometry with global shape context. Our key idea is to predict medial radii from estimated surface anchors and their geometric context. Queries sharing the same anchor and context receive the same radius, reflecting the constant radius along a medial spoke. Geometric constraints guide training without ground-truth medial axes or radii,
while consistency across input subsamples reduces sensitivity
to sampling. The predicted fields define a quasi-medial
distance field, from which we extract a medial mesh using
an existing shell extraction and collapse
procedure~\cite{kong2026qmdf}. Trained across shapes, the
network predicts fields for unseen inputs without test-time
network fitting.

Our contributions are:
\begin{itemize}
    \item A feed-forward network for predicting continuous
    signed distance and medial fields from sparse oriented
    point clouds, without per-shape network optimization.

    \item A medial branch that predicts radii from surface
    anchors and local--global shape context, sharing
    predictions across queries with the same boundary
    correspondence.

    \item A training strategy that combines geometric
    constraints with consistency across input subsamples,
    enabling medial prediction without ground-truth
    medial axes or radii.
\end{itemize}

We will publicly release our code and trained model weights
upon publication.

%% file: sections/related.tex
\section{Related Work}
\label{sec:related}

\subsection{Classical Medial Geometry}

Classical methods recover medial structure directly from a shape's boundary or volume~\cite{saha2016survey,tagliasacchi2016star}. Geometric constructions work with polyhedral boundaries~\cite{culver1999accurate,culver2004exact,milenkovic1993robust,sherbrooke1996algorithm} or trace medial structure from free-form surfaces~\cite{ramanathan2010interior}. For voxelized shapes, thinning methods progressively remove boundary voxels while preserving topology~\cite{palagyi1999parallel,pudney1998distance}. Other approaches locate skeletal points using nearest-boundary information~\cite{hesselink2008euclidean} or detect where the boundary distance function loses smoothness~\cite{rumpf2002continuous}. Voxel Cores uses Voronoi geometry to approximate the medial axis of a voxelized shape; under suitable sampling conditions, the approximation preserves the shape's topology and approaches the medial axis as sampling is refined~\cite{yan2018voxel}.

For point clouds, medial spheres---maximal spheres inscribed in the shape---can be estimated from surface positions and normals~\cite{ma2012medial,jalba2013surface}, or recovered through Voronoi constructions~\cite{brandt1994convergence,dey2004approximating}. Power Crust selects Voronoi balls to reconstruct both surface and medial structure~\cite{amenta2001power}. However, small changes in the sampled boundary can introduce many medial branches. This sensitivity motivates methods that remove unstable detail~\cite{attali1996modeling,chazal2005lambda}, preserve topology~\cite{giesen2006medial,chazal2008smooth}, or retain structure at a chosen scale~\cite{miklos2010discrete}. Beyond removing unstable branches, simplification methods reduce the number of medial spheres needed to represent the shape, using reconstruction error~\cite{li2015qmat,pan2019qmatplus} or surface coverage~\cite{dou2022coverage,wang2024coverage} to guide selection. When these spheres are connected into a medial mesh, further challenges include preserving sharp features~\cite{wang2022matfp}, maintaining topology~\cite{wang2024mattopo}, and improving mesh quality~\cite{wang2025matstruct}.

Medial Skeletal Diagrams further reduce representation size by allowing each skeletal element to capture more complex surface geometry~\cite{guo2024msd}. When the goal is to describe a shape's branching structure, curve-skeleton methods summarize it as a network of curves~\cite{tagliasacchi2009rosa,tagliasacchi2012mcf,huang2013l1}. Deep Points Consolidation combines curves and sheets to describe both tubular and flatter regions while jointly refining surface samples~\cite{wu2015dpc}. Our approach learns the relation between surface samples and medial geometry across shapes, predicting continuous medial fields for previously unseen point clouds.

\subsection{Neural Distance Fields}

Neural distance fields represent surfaces by predicting the distance from spatial queries to the boundary. Periodic activations and Fourier features help these networks capture fine geometric detail~\cite{sitzmann2020siren,tancik2020fourier}, while geometric constraints guide learning from surface observations. Eikonal regularization encourages the unit gradient norm of a distance function~\cite{gropp2020igr}, and several methods enable reconstruction from raw or unoriented points~\cite{atzmon2020sal,ma2021neuralpull,wang2024neuralimls}. Matching the observed surface alone, however, can leave distances elsewhere inaccurate. Further work addresses this through derivative-based regularization~\cite{benshabat2022digs,wang2023nsh}, more stable optimization~\cite{yang2023steik}, objectives based on phase transitions, heat flow, or viscosity~\cite{lipman2021phase,wang2025hotspot,krishnan2025viscoreg}, and bounds on how rapidly the predicted field can vary~\cite{coiffier2024lipschitz}. In our setting, the signed distance function (SDF) provides geometric constraints for learning medial radii, making its accuracy throughout the interior important for medial prediction.

Unsigned distance fields support open surfaces without requiring an inside--outside distinction~\cite{chibane2020ndf,ren2023geoudf,zhou2022capudf,fainstein2024dudf}. OffsetAxis connects this representation to medial geometry by recovering the surface as the medial axis of a thin surrounding volume~\cite{huang2026offsetaxis}.

\subsection{Learning Implicit Fields Across Shapes}

Learning from collections of shapes allows a model to reuse geometric knowledge when reconstructing a new object. Networks condition field queries on global shape codes~\cite{mescheder2019occnet,chen2019imnet,park2019deepsdf}, sets of latent vectors~\cite{zhang2023vecset}, or features stored on spatial grids and octrees~\cite{peng2020convonet,chibane2020ifnet,wang2022dog}. Some of these models still require optimization for each new shape. DeepSDF optimizes a global latent code~\cite{park2019deepsdf}, while local implicit grids and DeepLS optimize codes for individual spatial regions~\cite{jiang2020lig,chabra2020deepls}. Other methods use learned priors to guide per-shape SDF fitting or adaptation~\cite{ma2022osp,chen2024prior}, and neural kernel methods solve for reconstruction coefficients using predicted features~\cite{williams2022nkf,huang2023nksr}. GenSDF further examines generalization to categories absent from training~\cite{chou2022gensdf}.

Feed-forward methods instead predict fields directly from point clouds using learned geometric features~\cite{erler2020points2surf,boulch2022poco,erler2024ppsurf,li2025noksr}. A key design choice is how much of the input each query can access. Points2Surf combines local surface information for distance estimation with global shape information for inside--outside classification~\cite{erler2020points2surf}. We extend this local--global conditioning to medial prediction, where estimating the radius associated with a surface point requires relating its neighborhood to other regions of the boundary.

\subsection{Learning-based Medial Representations}

Learning-based methods such as P2MAT-NET, Point2Skeleton, and Point2MM predict medial spheres from point clouds, then connect their centers to form a skeleton~\cite{yang2020p2mat,lin2021point2skeleton,ge2023point2mm}. Deep Medial Voxels learns skeletal structure from medical imaging volumes~\cite{pepe2025dmv}, while Smart-Tree specializes medial prediction to tree point clouds~\cite{dobbs2023smarttree}. These internal representations can also guide surface reconstruction from images~\cite{tang2019skeleton} and shape generation~\cite{petrov2024gem3d}.

Other methods represent medial information through continuous fields. Neural Skeleton modifies SDF training to make skeletal structure easier to extract~\cite{clemot2023neural}. Deep Medial Fields learns the radius of the associated maximal inscribed sphere at each query~\cite{rebain2021deep}. Q-MDF subtracts the SDF magnitude from this radius field to obtain a field that vanishes on the medial axis~\cite{kong2026qmdf}. Medial-axis-aware SDF learning uses an auxiliary field to identify the axis and adjust regularization where the distance gradient changes abruptly~\cite{weidemaier2026maaware}. These methods fit a neural representation separately for each shape. We extend the medial-field and Q-MDF formulations~\cite{rebain2021deep,kong2026qmdf} to feed-forward prediction of medial fields. The resulting fields support medial-mesh extraction.

%% file: sections/background.tex
\section{Background}
\label{sec:background}

\begin{figure*}[!t]
    \centering
    \includegraphics[width=\linewidth]{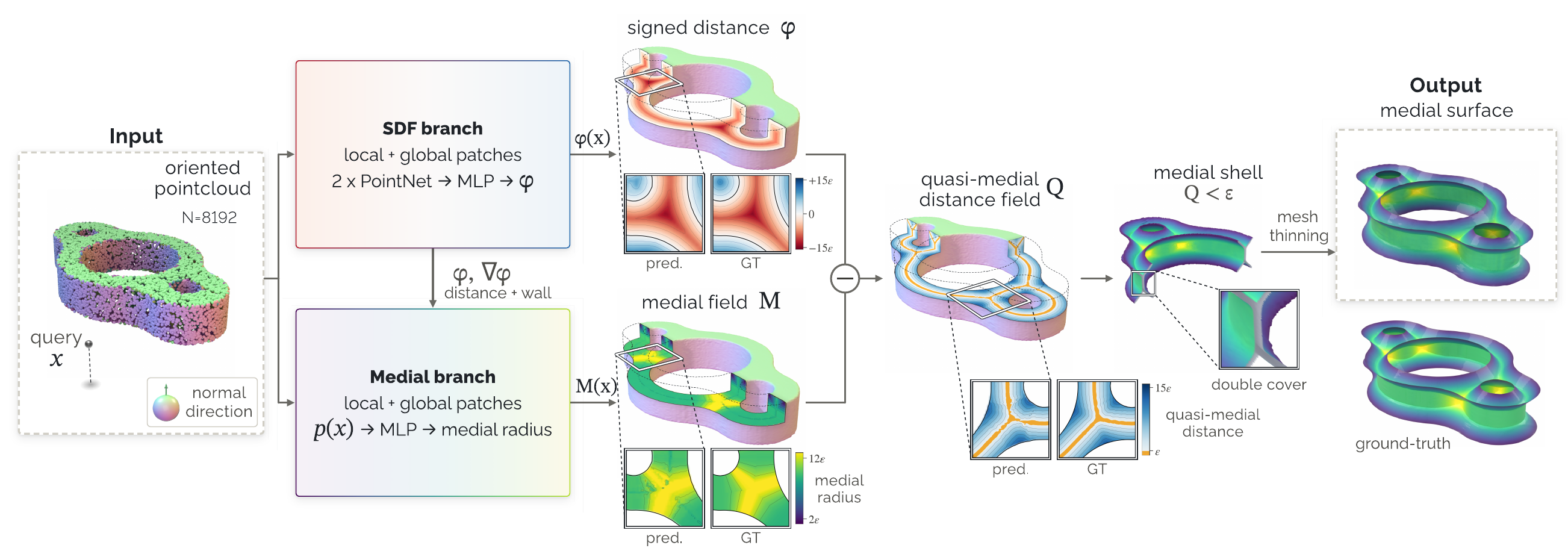}
    \caption{\textbf{Overview of Points2Medial.}
Given an oriented point cloud and a query $x$, the SDF branch predicts the signed distance $\phi(x)$. The medial branch uses $\phi$, $\nabla\phi$, and the input normals to estimate the corresponding surface point $p(x)$. It combines local geometry with global shape context to predict the radius $M(x)$ of the medial sphere tangent at that point. The resulting quasi-medial distance field $Q=M-|\phi|$ indicates the predicted medial axis through its near-zero values. Marching cubes extracts the level set $Q=\varepsilon$, where $\varepsilon$ is a small positive threshold controlling the thickness of the surrounding shell. This shell covers both sides of the medial sheets and is collapsed by mesh thinning into the final medial mesh. Insets compare predicted fields with ground truth.}
\label{fig:overview}
    \label{fig:architecture}
\end{figure*}

\subsection{Problem Setup}

Let $\Omega\subset\mathbb{R}^3$ be a bounded solid with watertight boundary $\partial\Omega$, interior $\Omega^{-}$, and exterior $\Omega^{+}$. Its interior medial axis $\mathcal{M}$ consists of the centers of maximal inscribed spheres~\cite{blum1967transformation,tagliasacchi2016star}. Given an oriented point cloud $\mathcal{P}=\{(p_i,n_i)\}_{i=1}^{N}$, where $p_i$ samples the boundary and $n_i$ is the outward unit normal, our goal is to recover $\mathcal{M}$ and its associated radii.

\subsection{Signed Distance and Medial Geometry}

The signed distance field $\phi$ is negative inside the shape, positive outside, and zero on the boundary. Its magnitude is the distance to the closest boundary point:
\begin{equation*}
    |\phi(x)|=\min_{p\in\partial\Omega}\|x-p\|.
\end{equation*}
Where $\phi$ is differentiable, it satisfies the eikonal equation $\|\nabla\phi(x)\|=1$, and the closest boundary point is
\begin{equation}
    p(x)=x-\phi(x)\nabla\phi(x).
    \label{eq:surface_point}
\end{equation}

Each medial center $c\in\mathcal{M}$ has radius $r(c)=|\phi(c)|$. The centers and radii form the medial axis transform~\cite{blum1967transformation}:
\begin{equation*}
    \mathrm{MAT}(\Omega)=\{(c,r(c)):c\in\mathcal{M}\}.
\end{equation*}
The corresponding balls cover the shape's interior, making the MAT a complete representation of the solid~\cite{amenta2001power}.

\subsection{Medial Field}

For an interior point $x\notin\mathcal{M}$, moving inward along $-\nabla\phi(x)$ first reaches a medial center $c(x)$. The segment from $p(x)$ to $c(x)$ is called a medial spoke~\cite{pizer2003mreps,damon2003skeletal}. The medial field assigns to $x$ the radius of its associated sphere~\cite{rebain2021deep}:
\begin{equation}
    M(x)=|\phi(c(x))|.
    \label{eq:medial_field}
\end{equation}

As $x$ moves along the spoke from the boundary to the center, its distance to the surface, $|\phi(x)|$, increases from zero to the sphere's radius. The medial field $M(x)$ remains equal to that radius throughout the segment. Therefore,
\begin{equation}
    M(x)\geq|\phi(x)|,
    \label{eq:maximality}
\end{equation}
with equality at the medial center, where $x=c(x)$. Since $M$ is constant along the spoke, its rate of change in the direction of $\nabla\phi$ is zero. Wherever both fields are differentiable off the axis,
\begin{equation}
    \nabla M(x)\cdot\nabla\phi(x)=0.
    \label{eq:ortho}
\end{equation}
The remaining distance along the spoke is $M(x)-|\phi(x)|$. Moving this distance inward along $-\nabla\phi(x)$ gives the associated center:
\begin{equation}
    c(x)=x-\big(M(x)-|\phi(x)|\big)\nabla\phi(x).
    \label{eq:medial_center}
\end{equation}

For an exact SDF, these conditions characterize the medial field under the assumptions of~\cite{rebain2021deep}. Deep Medial Fields turns them into training penalties. During learning, $\phi$ and $M$ denote neural predictions, and Eq.~\eqref{eq:medial_center} computes a candidate center $c(x)$ from them. For the interior setting considered here, the medial penalties are
\begin{align}
    \mathcal{L}_{\mathrm{max}}
    &=\int_{\Omega^{-}}
      \max\big(|\phi(x)|-M(x),0\big)^2\,dx,
    \label{eq:loss_maximality}\\
    \mathcal{L}_{\mathrm{ortho}}
    &=\int_{\Omega^{-}}
      \big(\nabla\phi(x)\cdot\nabla M(x)\big)^2\,dx,
    \label{eq:loss_ortho}\\
    \mathcal{L}_{\mathrm{ins}}
    &=\int_{\Omega^{-}}
      \big(|\phi(c(x))|-M(x)\big)^2\,dx.
    \label{eq:loss_inscription}
\end{align}
These integrals are approximated using sampled interior points and require no ground-truth medial centers or radii.

\subsection{Implicit Medial Axis}

The medial field $M$ stores sphere radii and therefore does not vanish on the medial axis. To extract the axis as a zero level set, we instead use the remaining spoke length, which becomes zero at the associated center. Following~\cite{kong2026qmdf}, we define the quasi-medial distance field (Q-MDF):
\begin{equation}
    Q(x)=M(x)-|\phi(x)|.
    \label{eq:qmdf}
\end{equation}
For exact fields inside the shape, $Q(x)=\|x-c(x)\|$. It is therefore nonnegative and vanishes precisely on $\mathcal{M}$, providing an implicit representation of the axis. The distance $Q(x)$ is measured along the spoke. Another point on the medial axis may be closer to $x$, so this is not always the shortest distance. Nevertheless, $Q(x)=0$ exactly when $x$ lies on the axis, which is the property needed for extraction.

%% file: sections/method.tex
\section{Method}
\label{sec:method}

We learn to predict the signed distance and medial fields introduced in Sec.~\ref{sec:background} from an oriented point cloud $\mathcal P$, without fitting a network to each new shape. Our key idea is to use the constant radius along a medial spoke to guide prediction: we estimate the surface point associated with each interior query and predict its medial radius from local and global shape geometry. As shown in Fig.~\ref{fig:architecture}, the SDF branch provides the distances used to locate these surface points, and the medial branch predicts their associated radii. The resulting fields $\hat\phi$ and $\hat M$ define $\hat Q$ through Eq.~\eqref{eq:qmdf}, from which we extract a medial mesh.

\subsection{Field Prediction}
\label{sec:architecture}

The SDF supplies the distances used in both the surface projection
and the quasi-medial field, so its accuracy throughout the interior
matters for medial prediction. For a query $x$, we start from its
distance to the nearest input point, $d(x)$, and learn a correction
from local and global surface features. The surface samples forming
the local descriptor may lie close to or far from the query.
We give the learned correction greater weight when nearby samples
are available and reduce its influence as the query moves farther
from the input cloud:
\begin{equation}
    \hat\phi(x)=\mathrm{sign}(s)\,
    \max\left(d(x)+\delta m\,e^{-\alpha d(x)},\,0\right).
    \label{eq:pred_sdf}
\end{equation}
The network predicts the scalar residual $m$ and the inside--outside
sign logit $s$. The fixed factor $\delta e^{-\alpha d(x)}$ controls
the contribution of the residual, with $\delta=0.01$ and
$\alpha=(3\delta)^{-1}$.
To predict $m$ and $s$, separate PointNet
encoders~\cite{qi2017pointnet} process a local patch of 300 nearest
points and a global sample of 1,024 points, adopting the
local--global conditioning of Points2Surf~\cite{erler2020points2surf}.

To predict a medial radius, we first estimate the corresponding
surface point, or anchor $p(x)$. We can approximate $p(x)$ using
Eq.~\eqref{eq:surface_point}, but errors in the learned SDF gradient
can affect its location. We therefore estimate the projection
direction from the input normals. Among the 16 nearest input
points, the normal most aligned with $\nabla\hat\phi(x)$ identifies
the boundary side to use. The normalized weighted average of
normals on that side gives the outward direction $n(x)$, avoiding the mixing of opposing normals.
The anchor is then
\begin{equation}
    p(x)=x-\hat\phi(x)n(x).
    \label{eq:pred_anchor}
\end{equation}

The medial branch predicts the radius of the maximal inscribed
sphere tangent to the surface at the anchor. The direction $n(x)$
is used to locate $p(x)$. At this location, we estimate a new
surface normal $n_p$ from the 300 nearest surface points, which
form the local patch. 
Both the local patch and a uniform global sample of 1,024 points are expressed in the local frame at $p(x)$, through features that are invariant to rotations about $n_p$ (the exact features are given in Eq.~\eqref{eq:medial_features}). Local features describe the surface near the anchor, while global
features capture more distant boundary regions that constrain the
size of the medial sphere. Two pointwise encoders, each followed
by max pooling, summarize the local and global point sets.
A decoder combines these features to predict a positive medial
radius $\hat M(x)$.

When optimizing, we omit the explicit orthogonality penalty used
in~\cite{rebain2021deep,kong2026qmdf}, since queries sharing a
surface anchor and its local frame receive the same radius.
With exact projections, this makes the radius constant along
each spoke and satisfies Eq.~\eqref{eq:ortho} by construction,
avoiding the higher-order derivatives required by the penalty.
Applying Eq.~\eqref{eq:qmdf} to the predicted fields gives
$\hat Q(x)=\hat M(x)-|\hat\phi(x)|$ inside the shape.
We set $\hat M=1$ for exterior queries, as reconstruction targets
only the inner medial axis.
\subsection{Training Objectives}
\label{sec:training}

The branches are trained sequentially: first the SDF branch, then the medial branch, whose anchors depend on its predictions. The SDF branch uses exact distances from the training meshes, with smooth-L1 distance supervision and binary cross-entropy for the sign. For the medial branch, we adapt the constraints described in Sec.~\ref{sec:background}~\cite{rebain2021deep} and add a consistency term across input samples. No medial-axis or radius annotations are required.

The maximality condition in Eq.~\eqref{eq:maximality} requires the radius to be at least the query's depth. Using the predicted radius, distance, and projection direction, we compute the candidate center as
\begin{equation}
    c(x)=x-\max\big(0,\hat M(x)-|\hat\phi(x)|\big)n(x).
    \label{eq:pred_medial_center}
\end{equation}
The clamp prevents the center from moving toward the boundary when the radius is smaller than the predicted depth. For inscription, the target must account for the fact that anchors are computed from the learned SDF. We therefore adjust the true distance at the candidate center by the SDF error at $x$:
\begin{equation}
    \tilde d(c(x))=|\phi(c(x))|
    +\big(|\hat\phi(x)|-|\phi(x)|\big).
    \label{eq:corrected_depth}
\end{equation}
Here $\phi$ is the ground-truth SDF. This correction assumes that the distance error $|\hat\phi|-|\phi|$ is approximately constant along the spoke. The maximality and inscription losses are
\begin{align}
    \mathcal L_{\mathrm{max}}
    &=\max\big(0,\,|\hat\phi(x)|-\hat M(x)\big),
    \label{eq:train_maximality}\\
    \mathcal L_{\mathrm{ins}}
    &=\big(\tilde d(c(x))-\hat M(x)\big)^2.
    \label{eq:train_inscription}
\end{align}
We use an unsquared maximality penalty; inscription compares the radius with the corrected distance at the candidate center.

To reduce sensitivity to the particular input samples, we evaluate the medial branch twice at the same anchors using different input samples (see Sec.~\ref{sec:configuration}, \textbf{Input and encoders}). The resulting predictions $\hat M$ and $\hat M_2$ are compared through
\begin{equation}
    \mathcal L_{\mathrm{cons}}=
    \big(\hat M(x)-\hat M_2(x)\big)^2.
    \label{eq:train_consistency}
\end{equation}
The total medial loss, averaged over the batch, is
\begin{equation}
    \mathcal L=
    \lambda_{\mathrm{max}}\mathcal L_{\mathrm{max}}+
    \lambda_{\mathrm{ins}}\mathcal L_{\mathrm{ins}}+
    \lambda_{\mathrm{cons}}\mathcal L_{\mathrm{cons}},
    \label{eq:train_total}
\end{equation}
where $\lambda_{\mathrm{max}}$, $\lambda_{\mathrm{ins}}$, and
$\lambda_{\mathrm{cons}}$ are nonnegative weights controlling
the contribution of each term.

\subsection{Medial Sheet Extraction}
\label{sec:extraction}

We apply the extraction procedure of Kong et al.~\cite{kong2026qmdf} to the predicted quasi-medial field. As shown on the right of Fig.~\ref{fig:architecture}, the small positive level set $\hat Q=\varepsilon$ forms a thin shell around the medial sheets. The shell interior is tetrahedralized, and the mesh is contracted by minimizing the mean of $\hat Q$ with volume and Laplacian regularization~\cite{hou2023dcudf}. Network weights remain fixed during this mesh optimization. The extraction settings are given below.

\subsection{Experimental Configuration}
\label{sec:configuration}

\paragraph{Input and encoders.}
Each shape provides 8,192 surface points with outward normals,
randomly split into two 4,096-point subsets each time we evaluate
the consistency loss.
The SDF's local and global samples are centered at the query.
The local patch is normalized by its radius; global sampling
favors nearby points. A quaternion spatial
transformer~\cite{guerrero2018pcpnet} predicts a rotation from
the global sample and aligns both context sets before encoding,
reducing sensitivity to shape orientation. The PointNet encoders and max pooling produce two 512-dimensional codes. Their concatenation enters an MLP with widths $1024\to256\to128\to2$, with BatchNorm and ReLU in the hidden layers, producing $m$ and $s$.

\paragraph{Anchor normals and medial features.}
As described above, $n(x)$ combines normals on the
selected boundary side, weighted by inverse squared distance to $x$. At the anchor $p(x)$, we estimate $n_p$ from same-side normals among the 300 nearest input points, using Gaussian weights based on distance to the anchor. Both averages are normalized, and $n_p$ is oriented inward. Local features use this 300-point neighborhood, while global features use a uniform sample of 1,024 input points, redrawn at each training step and fixed at test time.

To describe the surface geometry relative to the anchor
while remaining invariant to rotations about $n_p$, we use
distances and normal alignments. For an input point $y$ with normal $n_y$, let $d=y-p$. We decompose this displacement into a signed offset from the anchor's tangent plane, $v=d\cdot n_p$, and a tangential component, $t=d-vn_p$, with unit direction $\hat t$. The five input features are
\begin{equation}
    f(y)=\left(\|d\|,\ v,\ \|t\|,\
    n_y\cdot n_p,\ n_y\cdot\hat t\right),
    \label{eq:medial_features}
\end{equation}
The first three describe $y$ relative to $p$: its distance
to the anchor, signed offset along $n_p$, and tangential
distance. The last two describe normal orientation:
$n_y\cdot n_p$ helps distinguish the anchor's side from
facing surfaces, while $n_y\cdot\hat t$ indicates whether
the normal tilts toward or away from the anchor along the
tangent plane, providing a cue about local bending.
Lengths are expressed in units of $0.1$, the unit of $\hat M$. 
Each point set is encoded by an MLP with widths $5\to64\to128\to256$ and max pooling, producing two 256-dimensional codes. 
An MLP ($512\to256\to128\to1$) maps the concatenated codes to the medial radius, followed by a softplus that keeps $\hat M$ positive.

\paragraph{Data and optimization.}
Both branches are trained on ABC CAD~\cite{koch2019abc}: the SDF branch uses 100,000 shapes and the medial branch the first 10,000 of them.
We limit the medial training set to fit our computational
budget. Both branches take about $0.10$\,s per step, but medial
training processes one shape versus eight for SDF training,
requiring roughly eight times more time per shape.
SDF optimization uses SGD with momentum 0.9 and cosine decay, with eight shapes and 32 queries per shape per step. Samples are randomly rotated as in~\cite{erler2020points2surf}. The three stages in Table~\ref{tab:training_config} total 330,000 steps, each initialized from the preceding stage.
During warm-up, we omit the exponential factor and replace
$d(x)$ with a softened nearest-point distance $d_\tau(x)$
to facilitate initial optimization:
\begin{equation*}
    |\hat\phi(x)|=\max\big(d_\tau(x)+\delta m,0\big).
\end{equation*}
We compute $d_\tau(x)$ as a soft minimum over the distances
to the 32 nearest input points, smoothing the transitions
between nearest neighbors. 

Medial training uses Adam with cosine decay for 150,000 steps,
with 512 interior queries from one shape per step.
The loss weights in Eq.~\eqref{eq:train_total} are set to
$\lambda_{\mathrm{max}}=\lambda_{\mathrm{ins}}=100$
and $\lambda_{\mathrm{cons}}=400$.
\begin{table}[t]
    \centering
    \small
    \setlength{\tabcolsep}{3pt}
    \begin{tabular}{lrrc}
        \hline
        Stage & Shapes & Steps & Initial LR\\
        \hline
        SDF warm-up & 10,000 & 100,000 & $10^{-2}$\\
        SDF transition & 10,000 & 30,000 & $2\times10^{-2}$\\
        SDF full training & 100,000 & 200,000 & $5\times10^{-3}$\\
        Medial training & 10,000 & 150,000 & $10^{-3}$\\
        \hline
    \end{tabular}
    \caption{\textbf{Training schedule.}
Number of training shapes, optimization steps, and initial
learning rate (LR) for each stage.}
    \label{tab:training_config}
\end{table}
Training takes 11 hours for the SDF branch (3, 1, and 7 hours for its three stages) and 5 hours for the medial branch, using a single NVIDIA A100 (80\,GB VRAM) GPU, 8 cores Intel Xeon Silver 4310 CPU, and 16\,GB of RAM.

\paragraph{Query sampling.}
\label{sec:sampling}
Each SDF step draws a new input cloud for each shape. Its 32 queries comprise 16 interior points, eight near-surface points generated with Gaussian noise, $\sigma\in\{0.01,0.05\}$, and eight uniform samples in $[-1,1]^3$. Half the budget is assigned to interior queries because these support medial prediction.
Each medial step uses a fixed input cloud for the step and draws its 512-query batch from 2,000 interior points with precomputed SDF values. Half the batch is sampled randomly. For the other half, we draw 512 additional candidates, evaluate $\hat M(x)-|\phi(x)|$, and retain the 256 lowest-scoring points to focus training near the current medial prediction.

\paragraph{Extraction settings.}
The predicted SDF is sampled once per shape on a $128^3$ grid and interpolated trilinearly; $\hat Q$ is evaluated on a $256^3$ grid. Marching cubes uses $\varepsilon=0.005$ times the largest bounding-box side. The shell interior is tetrahedralized using restricted Delaunay, and contraction uses 300 VectorAdam iterations~\cite{ling2022vectoradam}. No mesh cleanup is applied.

\begin{figure*}[!t]
    \centering
    \includegraphics[width=\linewidth]{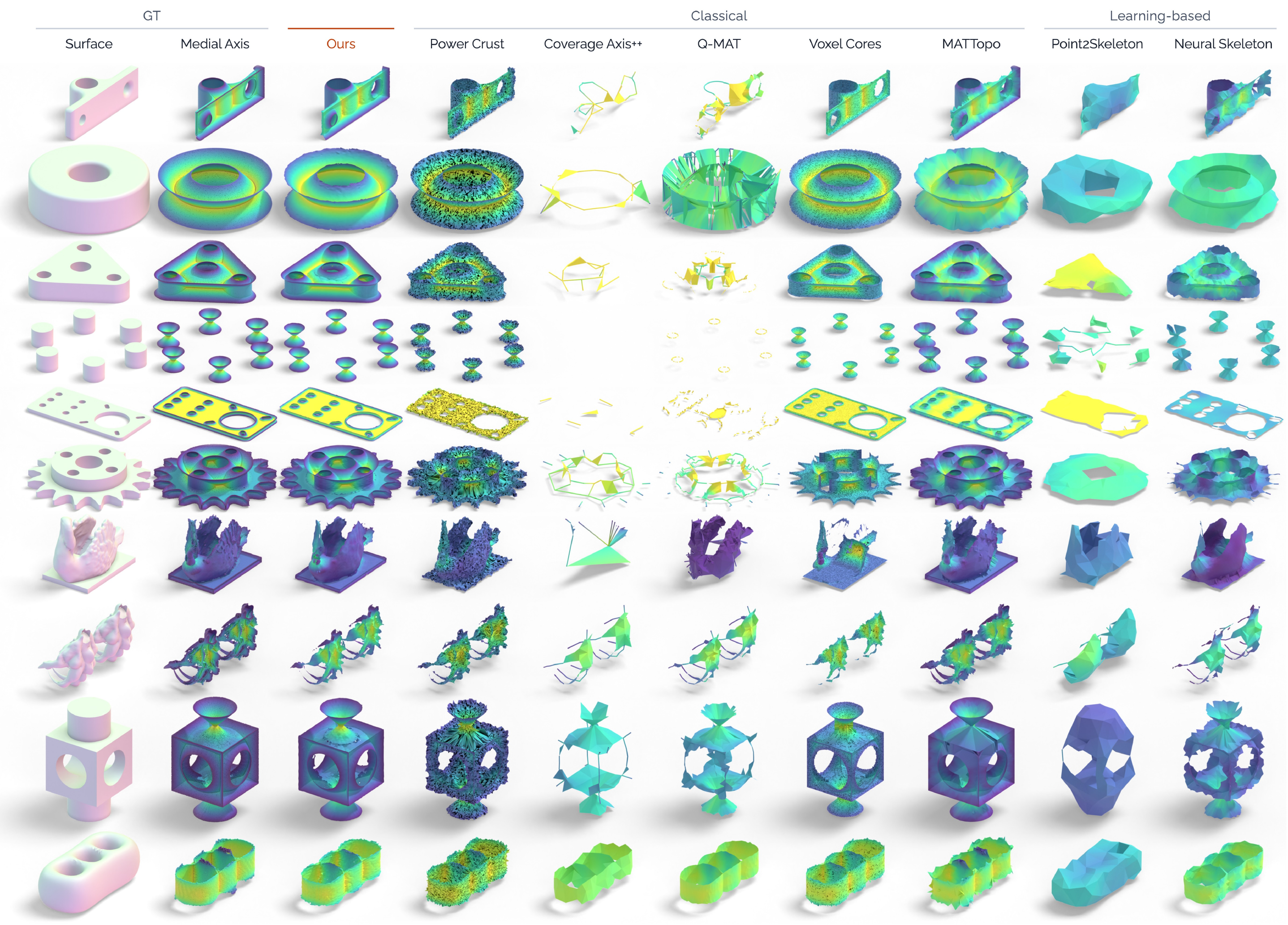}
    \caption{\textbf{Qualitative comparison of medial reconstructions.}
Each row shows a reference surface and its ground-truth
medial axis, followed by reconstructions from Points2Medial,
classical methods, and learning-based baselines.
The examples cover CAD and organic shapes with thin sheets,
holes, repeated components, and branching structures.
Points2Medial predicts its fields from 8,192-point oriented
clouds without per-shape network fitting, recovering the
main medial structures across these examples.}
    \label{fig:results}
\end{figure*}

\paragraph{Baselines.}
We compare against geometric methods Power Crust
(PC)~\cite{amenta2001power},
CoverageAxis++~\cite{wang2024coverage},
Q-MAT~\cite{li2015qmat},
Voxel Cores (VC)~\cite{yan2018voxel},
MATTopo~\cite{wang2024mattopo},
MATFP~\cite{wang2022matfp},
as well as learning-based methods
Point2Skeleton~\cite{lin2021point2skeleton} and
NeuralSkeleton~\cite{clemot2023neural}.
We also include Q-MDF~\cite{kong2026qmdf} to compare feed-forward prediction with per-shape
neural fitting. We use its tabulated results for selected
3D models, as its trained weights are unavailable and our
request to the authors went unanswered.

%% file: sections/results.tex
\section{Results}
\label{sec:results}

\paragraph{Qualitative comparison.}
Figure~\ref{fig:results} compares medial reconstructions on
CAD and organic shapes. Despite being trained only on ABC
CAD data, Points2Medial recovers the main medial sheets and
branching structures of organic shapes without fine-tuning.
This suggests that the learned local surface features
transfer across shape types, while global context supports
medial radius prediction beyond the training distribution.
The reconstructions also capture multiple disconnected
components (row~4), complex hole configurations (rows~3
and~9), and thin sheets (row~5).
In general, our outputs exhibit fewer visible irregularities than
Power Crust, while retaining details that CoverageAxis++,
Q-MAT, and Point2Skeleton simplify or omit in several
examples. Fine branches and sheet boundaries remain challenging
for our method, as discussed below.

\begin{table*}[!t]
    \centering
    \resizebox{\linewidth}{!}{%
\begin{tabular}{l cccccccccc}
\toprule
Shape & PC & CoverageAxis++ & Q-MAT & VC & MATTopo & Point2Skeleton & Neural Skeleton & \underline{MATFP} & \underline{Q-MDF} & Ours \\
\midrule
bunny    & 2.93 / 40.0 & 6.35 / 38.8   & 2.02 / 18.1 & 3.36 / 118.9 & \textbf{0.79} / 14.7         & 16.35 / 146.0 & 9.97 / 110.6 & 1.94 / 40.0         & 1.86 / \textbf{12.6} & 1.98 / 14.4 \\
kitten   & 1.05 / 27.9 & 3.60 / 34.7   & 1.50 / 18.7 & 2.80 / 83.2  & \textbf{0.76} / 57.2         & 17.23 / 163.6 & 10.96 / 38.3 & 0.79 / 24.4         & 1.58 / 20.0          & 1.30 / \textbf{11.6} \\
holes    & 0.83 / 12.1 & 3.04 / 25.9   & 0.59 / 6.0  & 2.13 / 7.7   & 0.75 / \textbf{5.8}          & 13.34 / 96.5  & 9.64 / 27.4  & 0.49 / 8.1          & 0.89 / 9.7           & \textbf{0.46} / 6.0 \\
trim-star & 2.68 / 53.6 & 4.93 / 61.4  & 1.67 / 24.3 & 2.56 / 22.6  & \textbf{0.56} / \textbf{8.8} & 13.32 / 80.4  & 5.70 / 36.5  & 0.99 / 15.9         & 0.86 / 16.5          & 1.32 / 14.1 \\
block    & 3.72 / 42.6 & 5.20 / 43.0   & 3.19 / 30.7 & 2.10 / 20.9  & 0.46 / 49.6                  & 15.02 / 111.0 & 6.27 / 44.2  & \textbf{0.32} / 9.6 & 0.95 / 21.4          & 1.07 / \textbf{8.3} \\
s-sphere & 5.77 / 78.6 & 11.86 / 100.5 & 4.10 / 54.2 & 3.93 / 52.5  & \textbf{0.59} / \textbf{7.3} & 16.96 / 96.9  & 9.10 / 60.4  & 0.68 / 16.5         & 1.67 / 19.7          & 1.56 / 30.5 \\
\midrule
mean     & 2.83 / 42.5 & 5.83 / 50.7   & 2.18 / 25.3 & 2.81 / 51.0  & \textbf{0.65} / 23.9         & 15.37 / 115.7 & 8.61 / 52.9  & 0.87 / 19.1         & 1.30 / 16.7          & 1.28 / \textbf{14.2} \\
\bottomrule
\end{tabular}}
\caption{\textbf{Surface reconstruction metrics.}
CD/HD ($\times 10^3$; lower is better) for surfaces reconstructed
from medial axis transforms on six shapes from
Q-MDF~\cite{kong2026qmdf}. Underlined results are taken from
that work because the implementations were unavailable or
could not be run successfully in our setup; we evaluated
the remaining baselines ourselves. Points2Medial predicts
fields from 8,192-point oriented clouds without per-shape
network fitting. The mean covers all six shapes, and the
lowest value for each metric in each row is shown in bold.}
    \label{tab:qmdf_shapes}
\end{table*}

\begin{table}[!t]
    \centering
    \setlength{\tabcolsep}{3pt}
    \begin{tabular*}{\columnwidth}{@{\extracolsep{\fill}}lcccc@{}}
        \toprule
        & \multicolumn{2}{c}{CD $\downarrow$}
        & \multicolumn{2}{c}{HD $\downarrow$} \\
        \cmidrule(lr){2-3}\cmidrule(lr){4-5}
        Dataset & Q-MDF & Ours & Q-MDF & Ours \\
        \midrule
        ABC
        & 1.60 & $\mathbf{1.56}\pm0.62$
        & 20.4 & $\mathbf{19.5}\pm10.8$ \\
        Thingi10K
        & 2.60 & $\mathbf{1.93}\pm0.74$
        & \textbf{18.9} & $20.0\pm7.5$ \\
        \bottomrule
    \end{tabular*}
    \caption{\textbf{Generalization across datasets.}
CD and HD ($\times 10^3$), with mean $\pm$ standard deviation
for ours. Trained only on ABC, Points2Medial uses 8,192
oriented points and is evaluated without fine-tuning on
100 ABC and the first 50 Thingi10K test shapes of
Points2Surf~\cite{erler2020points2surf}.
Q-MDF fits each shape separately; its published
means~\cite{kong2026qmdf} use different, unpublished subsets
of 30--50 shapes per dataset. Lowest reported means are
in bold.}
    \label{tab:abc_thingi}
\end{table}

\begin{table}[!t]
\centering
\small
\setlength{\tabcolsep}{3pt}
\begin{tabular*}{\columnwidth}{@{}l@{\extracolsep{\fill}}cccc@{}}
\toprule
 & CD$\downarrow$ & HD$\downarrow$ & Miss $\downarrow$ (\%) & M err $\downarrow$ \\
\midrule
Full model       & 1.79 & 21.6 & 8.4  & 0.05\,/\,0.12 \\
\midrule
w/o maximality   & 1.63 & 21.1 & 12.2 & 0.14\,/\,0.32 \\
w/o inscription  & --   & --   & --   & 102\,/\,125 \\
w/o consistency  & 1.83 & 23.0 & 6.7  & 0.04\,/\,0.11 \\
w/o residual SDF & 1.87 & 25.3 & 8.5  & 0.05\,/\,0.12 \\
\bottomrule
\end{tabular*}
\caption{\textbf{Component ablations.}
Evaluation of a smaller model on the first 30 ABC test
shapes; ``Full model'' retains all components.
CD and HD ($\times 10^3$) measure distances between predicted
and ground-truth medial axes, averaged over shapes.
Miss is the percentage of ground-truth medial points
farther than $2\varepsilon$ from the predicted axis.
$M$ err reports median $|M-r|/\varepsilon$ for thin
($r<4\varepsilon$) and thick ($r\geq4\varepsilon$) spheres,
respectively, where $r$ is the ground-truth medial radius.
Dashes indicate that no medial mesh was extracted.}
\label{tab:ablation}
\end{table}

\paragraph{Reconstruction accuracy.}
Table~\ref{tab:qmdf_shapes} quantifies the accuracy of surfaces
reconstructed from medial axis transforms on six shapes.
Points2Medial achieves the lowest mean HD among the compared
methods without per-shape network fitting. MATTopo and MATFP achieve lower mean CD. These results should be
interpreted in light of their input requirements:
MATTopo~\cite{wang2024mattopo} operates on a tetrahedral mesh,
while MATFP~\cite{wang2022matfp} uses a triangle surface mesh,
with both using pre-detected sharp features for CAD models.
Points2Medial achieves competitive reconstruction accuracy
from only oriented points.

\paragraph{Generalization across datasets.}
Table~\ref{tab:abc_thingi} reports mean CD/HD on 100 ABC shapes and 50 Thingi10K shapes.
Points2Medial achieves these results with a single model
trained only on ABC, without test-time network fitting
or fine-tuning on Thingi10K. Performance on Thingi10K
indicates that the learned representation transfers
beyond the training dataset.
The accuracy is comparable to Q-MDF's published results,
which require training a separate network for each
evaluated shape. 
The different shape subsets, however, limit direct comparison.

\paragraph{Ablations.}
We remove one component at a time, training each variant
on 1,000 ABC shapes and evaluating it on the first 30 ABC
test shapes (Table~\ref{tab:ablation}).
Removing maximality increases the missing fraction from
8.4\% to 12.2\%, despite a lower CD, indicating poorer
coverage of the medial axis.
Without inscription, median radius errors exceed
$100\varepsilon$ and no medial mesh is extracted.
Removing consistency increases HD from 21.6 to 23.0,
although the missing fraction decreases.
Removing the residual SDF formulation produces the largest
HD increase among variants that yield a mesh, reaching 25.3.

%% file: sections/limitations.tex
\label{sec:limitations}
\paragraph{Limitations and Future Work.}
Our input is limited to 8,192 surface points, which may leave
fine geometric details insufficiently sampled and prevent
their recovery in the medial structure. The method also
requires consistently oriented normals, and errors in the
normals or predicted SDF can affect anchor placement and
medial prediction. Extracted medial meshes may exhibit holes
and ragged boundaries. Although
field prediction is feed-forward, mesh extraction still
requires dense grid evaluation and iterative contraction.
Future work could explore adaptive sampling and larger
input clouds to capture finer details, learning from
unoriented inputs, and joint refinement of surface and
medial predictions. Improving the continuity and efficiency
of medial mesh extraction is another promising direction.

%% file: sections/conclusion.tex
\section{Conclusion}
\label{sec:conclusion}

We presented Points2Medial, a feed-forward model for predicting
signed distance and medial fields from sparse oriented point
clouds without test-time network fitting. Our method predicts
medial radii from estimated surface anchors and local--global
shape context, using geometric constraints without ground-truth
medial supervision. Trained only on ABC, it generalizes to
Thingi10K with competitive medial reconstruction accuracy.
Improved SDF estimation deep inside shapes is
an additional benefit of the approach.

%% file: sections/AI_statement.tex
\section{LLM Usage Statement}
We used ChatGPT (OpenAI) to assist with drafting and revising the manuscript, checking technical explanations and related-work references, formatting LaTeX tables, and developing preliminary figure designs. We used Claude (Anthropic) for coding assistance. The authors reviewed and revised the resulting material and retain full responsibility for the implementation, experiments, references, and final manuscript.

%% file: sections/acknowledgements.tex
\section{Acknowledgements}
We thank Universidad Torcuato Di Tella for its support and Mikhail Bessmeltsev for valuable discussions on the medial axis problem.

%% file: sections/supplementary.tex
\begin{figure*}[!t]
    \centering
    \includegraphics[width=\linewidth]{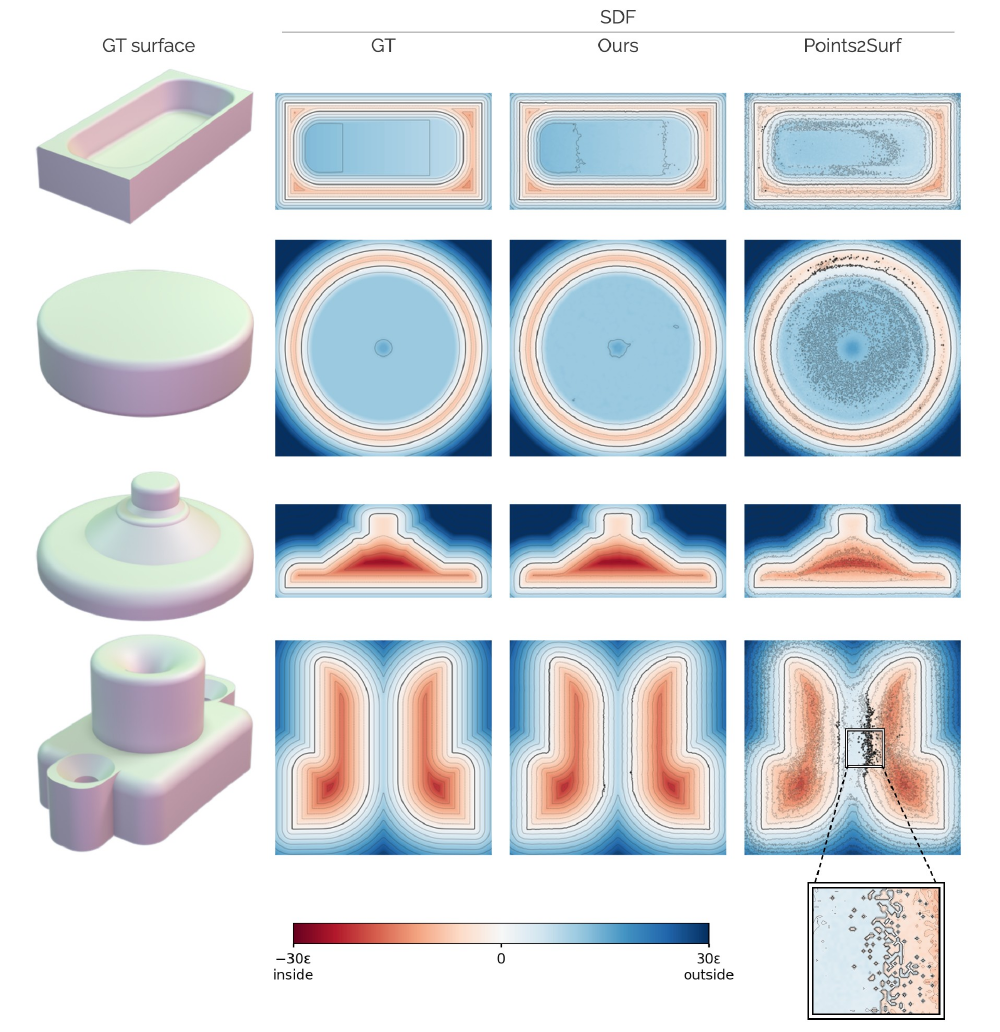}
    \caption{\textbf{Signed distance fields}. Ours vs.\ Points2Surf. Four ABC test shapes, one per row: the ground-truth surface,
    then a slice through the centre of the bounding box of the exact SDF (GT), our SDF (8,192-point oriented cloud) and the SDF of the released Points2Surf vanilla model. All cross sections share one colour
    scale.
    Our field closely matches the exact one, inside and outside the shape.
    The close-up shows the Points2Surf cross section of the last shape near a wall.
    Each small closed black loop there is a spurious zero-level island, where the predicted sign flips away from the surface.}
    \label{fig:sdf_comparison}
\end{figure*}

\twocolumn[{%
\section{Supplementary}

Table~\ref{tab:layers} lists the layers of both branches, with the input and output size of each module.
Figure~\ref{fig:ablations} shows the ablated variants of Table~\ref{tab:ablation} on two ABC test shapes.
Figure~\ref{fig:limitations} shows extended qualitative results on the best and worst shapes and discusses the most common failures.
Finally, we compare our SDF with that of Points2Surf. Figure~\ref{fig:sdf_comparison} shows cross sections of both fields next to the exact SDF on four ABC test shapes. Table~\ref{tab:sdf_depth} reports the median SDF error on the ABC and Famous (from~\cite{erler2020points2surf}) test
sets, binned by the distance to the surface, inside and outside the shape.

\vspace{1em}
\begin{center}
    \includegraphics[width=\linewidth]{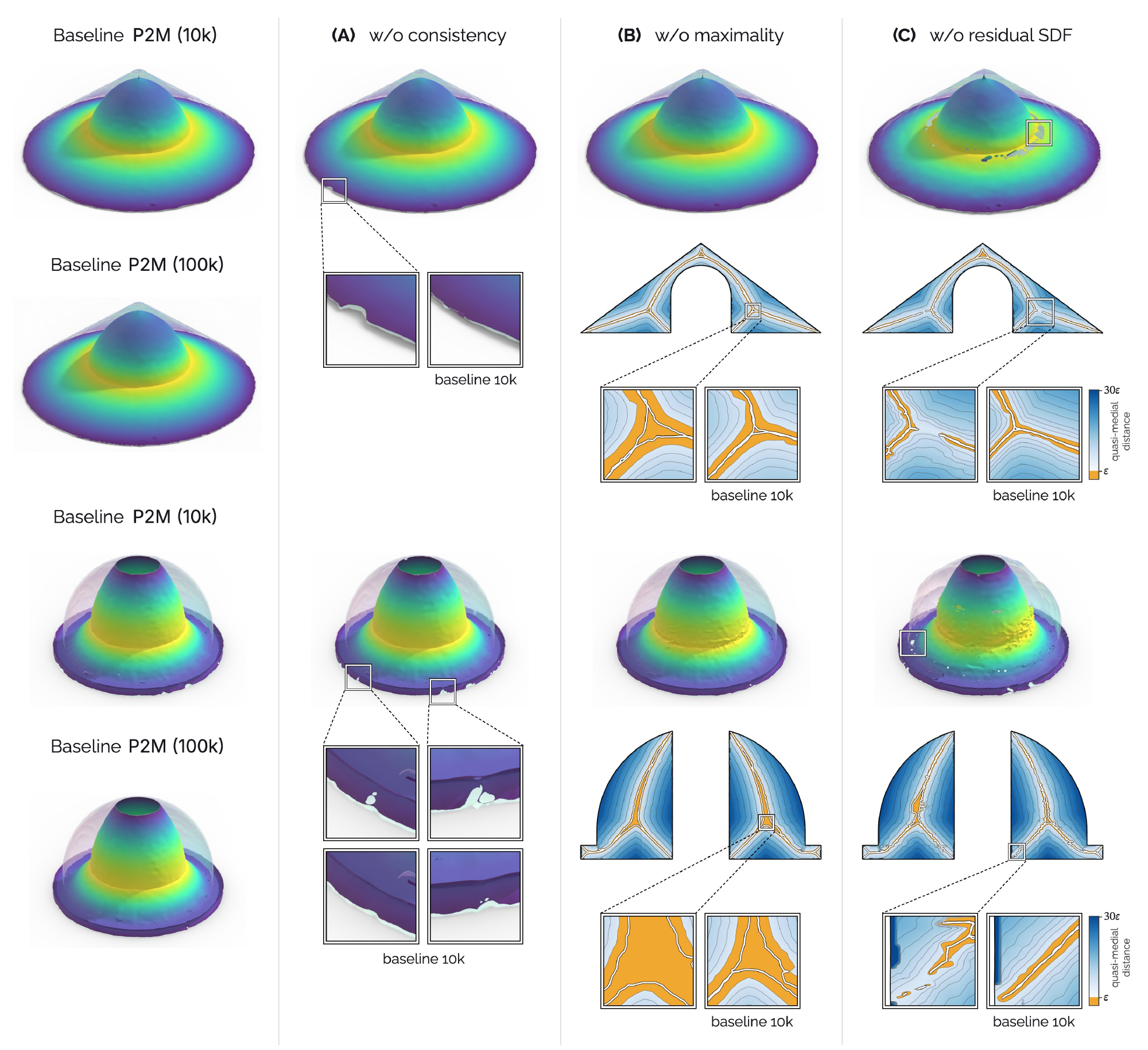}
    \captionof{figure}{\textbf{Qualitative ablations} on two ABC test shapes (Table~\ref{tab:ablation}).
    All variants are trained from the reduced model (P2M 10k); the full model (P2M 100k) gives all other results in the paper. Both recover clean,
    complete medial surfaces. Each remaining column removes one component. In the cross sections, the $\varepsilon$-shell is shown
    in orange and the thinned medial axis in white.
    \textbf{(A)} Without the consistency term (Eq.~\ref{eq:train_consistency}), predictions depend on the input sample,
    and the radius becomes unstable near the sheet boundaries, leaving ragged edges.
    \textbf{(B)} Without the maximality penalty (Eq.~\ref{eq:train_maximality}), radii may fall below the query depth.
    The surfaces look similar, but the cross sections show a wider $\varepsilon$-shell around junctions, which blurs the
    location of the axis.
    \textbf{(C)} Predicting the SDF directly, rather than as a residual on the nearest-point distance
    (Eq.~\ref{eq:pred_sdf}), makes distances less accurate. Since $\hat{Q} = \hat{M} - |\hat{\phi}|$, the
    $\varepsilon$-shell breaks into fragments, and the medial surfaces get holes and bumps, matching the largest HD
    increase in Table~\ref{tab:ablation}.}
    \label{fig:ablations}
\end{center}
}]

\twocolumn[{%
\begin{center}
\begin{minipage}{0.75\linewidth}
    \centering
    \footnotesize
    \setlength{\tabcolsep}{12pt}
    \begin{tabular}{@{}llll@{}}
    \toprule
    Module & Input & Layers & Output \\
    \midrule
    \multicolumn{4}{@{}l}{\emph{SDF branch}} \\
    Global enc. & [1024, 3] & QSTN + PointNet + FC & 512 \\
    Local enc.  & [300, 3] & PointNet + FC & 512 \\
    Head        & 1024 & FC(1024,256,128,2) & $m$, $s$ \\
    Distance    & $m$, $s$, $d(x)$ & Eq.~\ref{eq:pred_sdf} & $\hat\phi(x)$ \\
    \midrule
    \multicolumn{4}{@{}l}{\emph{M branch}} \\
    Surface anchor & $x$, $\hat\phi(x)$ & Eq.~\ref{eq:pred_anchor} & $p(x)$ \\
    Local enc.  & [300, 5] & MLP(5,64,128,256) + max & 256 \\
    Global enc. & [1024, 5] & MLP(5,64,128,256) + max & 256 \\
    Head        & 512 & MLP(512,256,128,1) & $\hat M(x)$ \\
    \bottomrule
    \end{tabular}
    \captionof{table}{\textbf{Layers of the two branches}. Query point $x$ and
    input cloud $P$ (8192 oriented points).
    PointNet: MLP(3,64,64,64,128,1024) + max pooling.
    SDF branch: 1024 points sampled around $x$ and the 300 nearest $x$, as coordinates.
    M branch: 1024 uniform points and the 300 nearest $p$, as the 5 features of Eq.~\ref{eq:medial_features}; a softplus keeps $\hat M$ positive.
    Parameters: 6.2\,M (SDF branch) and 0.25\,M (M branch).}
    \label{tab:layers}
\end{minipage}
\end{center}
\vspace{0.5em}

\begin{center}
    \includegraphics[width=\linewidth]{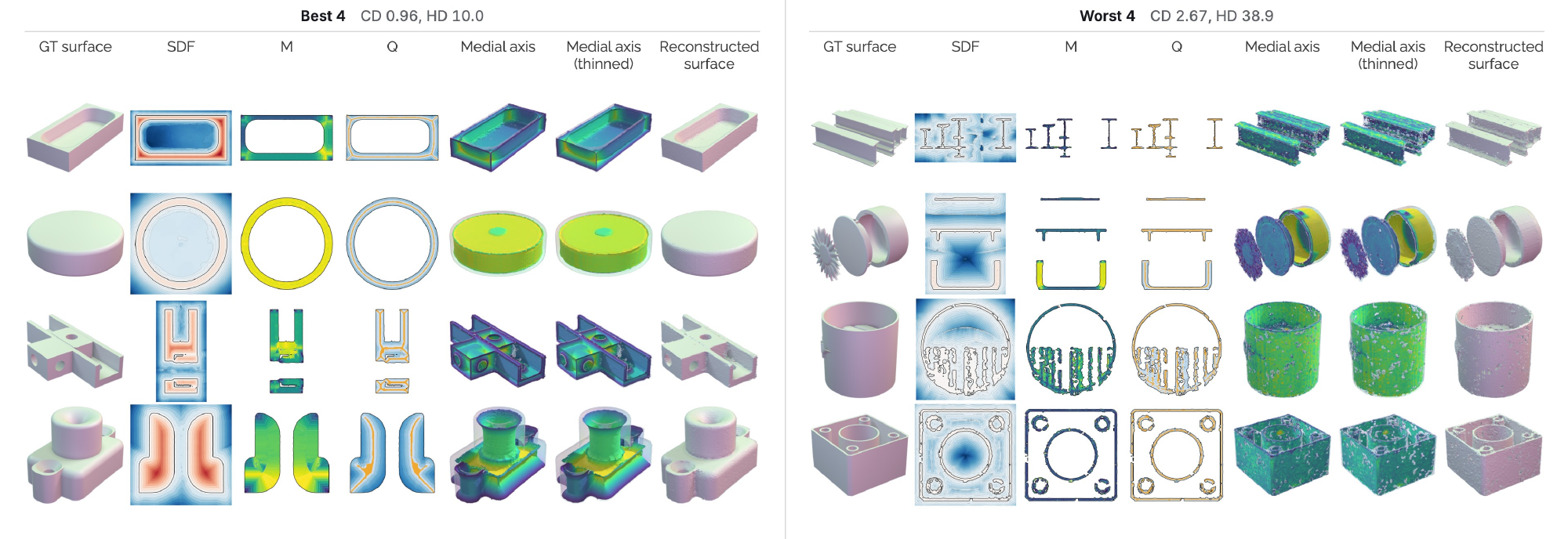}
    \captionof{figure}{\textbf{Limitations}. Best and worst cases on the ABC test set. Left: four of the best shapes (CD $0.96 \pm 0.13$, HD $10 \pm 3.6$). Right: four of the worst shapes (CD $2.67 \pm 0.25$, HD $38.9 \pm 11.8$), ranked by the sum of their CD and HD ranks among the 100 test shapes.
    Each row shows the ground-truth surface, cross section of the predicted SDF $\hat{\phi}$, medial field $\hat{M}$ and quasi-medial field $\hat{Q}$, the extracted medial axis before and after thinning, and the surface reconstructed from the medial axis transform.
    On the worst shapes, the predicted fields become noisy, the $\varepsilon$-shell of $\hat{Q}$ fragments, and the extracted medial sheets are bumpy, ragged, or contain holes and spurious pieces; the large HD values come mostly from such extra sheet fragments rather than from missing parts.
    Errors in the predicted SDF near close, opposing walls also misplace the
    surface anchors, which corrupts $\hat{M}$.}
    \label{fig:limitations}
\end{center}
\vspace{0.5em}

\begin{center}
\footnotesize
\setlength{\tabcolsep}{5pt}
\begin{tabular}{@{}ll cccc cccc@{}}
\toprule
 & & \multicolumn{4}{c}{Inside} & \multicolumn{4}{c}{Outside} \\
\cmidrule(lr){3-6}\cmidrule(lr){7-10}
Dataset & Method & $>10\varepsilon$ & $3$--$10\varepsilon$ & $1$--$3\varepsilon$ & $<1\varepsilon$ & $<1\varepsilon$ & $1$--$3\varepsilon$ & $3$--$10\varepsilon$ & $>10\varepsilon$ \\
\midrule
\multirow{3}{*}{ABC}    & Points2Surf (max)     & 5.26 & 1.60 & 0.37 & 0.32 & 0.41 & 0.41 & 0.86 & 2.50 \\
                        & Points2Surf (vanilla) & 2.60 & 0.85 & 0.29 & 0.30 & 0.30 & 0.35 & 0.65 & 2.56 \\
                        & Ours                  & \textbf{0.08} & \textbf{0.07} & \textbf{0.08} & \textbf{0.11} & \textbf{0.13} & \textbf{0.10} & \textbf{0.08} & \textbf{0.09} \\
\midrule
\multirow{3}{*}{Famous} & Points2Surf (max)     & 6.62 & 1.42 & 0.40 & 0.45 & 0.59 & 0.57 & 0.83 & 4.39 \\
                        & Points2Surf (vanilla) & 3.32 & 0.70 & 0.35 & 0.43 & 0.47 & 0.49 & 0.77 & 4.29 \\
                        & Ours                  & \textbf{0.08} & \textbf{0.09} & \textbf{0.11} & \textbf{0.12} & \textbf{0.18} & \textbf{0.14} & \textbf{0.11} & \textbf{0.13} \\
\bottomrule
\end{tabular}
\captionof{table}{\textbf{SDF accuracy by depth}. Median absolute SDF error $|\hat{\phi} - \phi|$ (in $\varepsilon$, 0.5\% of the
bounding box) on 100 ABC and 16 watertight Famous test shapes from~\cite{erler2020points2surf}, binned by the ground-truth signed distance.
Ours uses the 8,192-point oriented cloud; Points2Surf uses its released pretrained models on the full scan, queried at the
same points. Points2Surf's error grows away from the surface, while ours stays below $0.2\varepsilon$ at every depth.
The Points2Surf cross sections in Fig.~\ref{fig:sdf_comparison} use the vanilla model, the more accurate of the two here.}
\label{tab:sdf_depth}
\end{center}
}]

%% file: main.bib
@String(CVPR= {IEEE Conf. Comput. Vis. Pattern Recog.})

@String(ICCV= {Int. Conf. Comput. Vis.})

@String(ECCV= {Eur. Conf. Comput. Vis.})

@String(ICIP = {IEEE Int. Conf. Image Process.})

@String(CVPR  = {CVPR})

@String(ICCV  = {ICCV})

@String(ECCV  = {ECCV})

@String(ICIP  = {ICIP})

@inproceedings{sitzmann2020siren,
  author    = {Sitzmann, Vincent and Martel, Julien and Bergman, Alexander and Lindell, David and Wetzstein, Gordon},
  title     = {Implicit Neural Representations with Periodic Activation Functions},
  booktitle = {Advances in Neural Information Processing Systems},
  volume    = {33},
  pages     = {7462--7473},
  year      = {2020}
}

@inproceedings{tancik2020fourier,
  author    = {Tancik, Matthew and Srinivasan, Pratul and Mildenhall, Ben and Fridovich-Keil, Sara and Raghavan, Nithin and Singhal, Utkarsh and Ramamoorthi, Ravi and Barron, Jonathan and Ng, Ren},
  title     = {Fourier Features Let Networks Learn High Frequency Functions in Low Dimensional Domains},
  booktitle = {Advances in Neural Information Processing Systems},
  volume    = {33},
  pages     = {7537--7547},
  year      = {2020}
}

@inproceedings{gropp2020igr,
  author    = {Gropp, Amos and Yariv, Lior and Haim, Niv and Atzmon, Matan and Lipman, Yaron},
  title     = {Implicit Geometric Regularization for Learning Shapes},
  booktitle = {Proceedings of the 37th International Conference on Machine Learning},
  series    = {Proceedings of Machine Learning Research},
  volume    = {119},
  pages     = {3789--3799},
  publisher = {PMLR},
  year      = {2020}
}

@inproceedings{atzmon2020sal,
  author    = {Atzmon, Matan and Lipman, Yaron},
  title     = {{SAL}: Sign Agnostic Learning of Shapes From Raw Data},
  booktitle = {Proceedings of the IEEE/CVF Conference on Computer Vision and Pattern Recognition (CVPR)},
  year      = {2020}
}

@inproceedings{ma2021neuralpull,
  author    = {Ma, Baorui and Han, Zhizhong and Liu, Yu-Shen and Zwicker, Matthias},
  title     = {Neural-Pull: Learning Signed Distance Function from Point Clouds by Learning to Pull Space onto Surface},
  booktitle = {Proceedings of the 38th International Conference on Machine Learning},
  series    = {Proceedings of Machine Learning Research},
  volume    = {139},
  pages     = {7246--7257},
  publisher = {PMLR},
  year      = {2021}
}

@article{wang2024neuralimls,
  author    = {Wang, Zixiong and Wang, Pengfei and Wang, Pengshuai and Dong, Qiujie and Gao, Junjie and Chen, Shuangmin and Xin, Shiqing and Tu, Changhe and Wang, Wenping},
  title     = {Neural-{IMLS}: Self-supervised Implicit Moving Least-Squares Network for Surface Reconstruction},
  journal   = {IEEE Transactions on Visualization and Computer Graphics},
  volume    = {30},
  number    = {8},
  pages     = {5018--5033},
  year      = {2024},
  doi       = {10.1109/TVCG.2023.3284233}
}

@inproceedings{benshabat2022digs,
  author    = {Ben-Shabat, Yizhak and Koneputugodage, Chamin Hewa and Gould, Stephen},
  title     = {{DiGS}: Divergence Guided Shape Implicit Neural Representation for Unoriented Point Clouds},
  booktitle = {Proceedings of the IEEE/CVF Conference on Computer Vision and Pattern Recognition (CVPR)},
  pages     = {19323--19332},
  year      = {2022}
}

@inproceedings{yang2023steik,
  author    = {Yang, Huizong and Sun, Yuxin and Sundaramoorthi, Ganesh and Yezzi, Anthony},
  title     = {{StEik}: Stabilizing the Optimization of Neural Signed Distance Functions and Finer Shape Representation},
  booktitle = {Advances in Neural Information Processing Systems},
  volume    = {36},
  pages     = {13993--14004},
  year      = {2023}
}

@article{wang2023nsh,
  author    = {Wang, Zixiong and Zhang, Yunxiao and Xu, Rui and Zhang, Fan and Wang, Peng-Shuai and Chen, Shuangmin and Xin, Shiqing and Wang, Wenping and Tu, Changhe},
  title     = {Neural-Singular-Hessian: Implicit Neural Representation of Unoriented Point Clouds by Enforcing Singular Hessian},
  journal   = {ACM Transactions on Graphics},
  volume    = {42},
  number    = {6},
  year      = {2023},
  doi       = {10.1145/3618311}
}

@inproceedings{lipman2021phase,
  author    = {Lipman, Yaron},
  title     = {Phase Transitions, Distance Functions, and Implicit Neural Representations},
  booktitle = {Proceedings of the 38th International Conference on Machine Learning},
  series    = {Proceedings of Machine Learning Research},
  volume    = {139},
  pages     = {6702--6712},
  publisher = {PMLR},
  year      = {2021}
}

@inproceedings{wang2025hotspot,
  author    = {Wang, Zimo and Wang, Cheng and Yoshino, Taiki and Tao, Sirui and Fu, Ziyang and Li, Tzu-Mao},
  title     = {{HotSpot}: Signed Distance Function Optimization with an Asymptotically Sufficient Condition},
  booktitle = {Proceedings of the Computer Vision and Pattern Recognition Conference (CVPR)},
  pages     = {1276--1286},
  year      = {2025}
}

@misc{krishnan2025viscoreg,
  author        = {Krishnan, Meenakshi and Duraiswami, Ramani},
  title         = {{ViscoReg}: Neural Signed Distance Functions via Viscosity Solutions},
  year          = {2025},
  eprint        = {2507.00412},
  archivePrefix = {arXiv},
  primaryClass  = {cs.GR}
}

@article{coiffier2024lipschitz,
  author    = {Coiffier, Guillaume and B{\'e}thune, Louis},
  title     = {1-{L}ipschitz Neural Distance Fields},
  journal   = {Computer Graphics Forum},
  volume    = {43},
  number    = {5},
  pages     = {e15128},
  year      = {2024},
  doi       = {10.1111/cgf.15128}
}

@inproceedings{chibane2020ndf,
  author    = {Chibane, Julian and Mir, Aymen and Pons-Moll, Gerard},
  title     = {Neural Unsigned Distance Fields for Implicit Function Learning},
  booktitle = {Advances in Neural Information Processing Systems},
  volume    = {33},
  pages     = {21638--21652},
  year      = {2020}
}

@inproceedings{ren2023geoudf,
  author    = {Ren, Siyu and Hou, Junhui and Chen, Xiaodong and He, Ying and Wang, Wenping},
  title     = {{GeoUDF}: Surface Reconstruction from {3D} Point Clouds via Geometry-guided Distance Representation},
  booktitle = {Proceedings of the IEEE/CVF International Conference on Computer Vision (ICCV)},
  pages     = {14214--14224},
  year      = {2023}
}

@inproceedings{zhou2022capudf,
  author    = {Zhou, Junsheng and Ma, Baorui and Liu, Yu-Shen and Fang, Yi and Han, Zhizhong},
  title     = {Learning Consistency-Aware Unsigned Distance Functions Progressively from Raw Point Clouds},
  booktitle = {Advances in Neural Information Processing Systems},
  volume    = {35},
  pages     = {16481--16494},
  year      = {2022}
}

@inproceedings{fainstein2024dudf,
  author    = {Fainstein, Miguel and Siless, Viviana and Iarussi, Emmanuel},
  title     = {{DUDF}: Differentiable Unsigned Distance Fields with Hyperbolic Scaling},
  booktitle = {Proceedings of the IEEE/CVF Conference on Computer Vision and Pattern Recognition (CVPR)},
  pages     = {4484--4493},
  year      = {2024}
}

@inproceedings{mescheder2019occnet,
  author    = {Mescheder, Lars and Oechsle, Michael and Niemeyer, Michael and Nowozin, Sebastian and Geiger, Andreas},
  title     = {Occupancy Networks: Learning {3D} Reconstruction in Function Space},
  booktitle = {Proceedings of the IEEE/CVF Conference on Computer Vision and Pattern Recognition (CVPR)},
  year      = {2019}
}

@inproceedings{chen2019imnet,
  author    = {Chen, Zhiqin and Zhang, Hao},
  title     = {Learning Implicit Fields for Generative Shape Modeling},
  booktitle = {Proceedings of the IEEE/CVF Conference on Computer Vision and Pattern Recognition (CVPR)},
  year      = {2019}
}

@inproceedings{park2019deepsdf,
  author    = {Park, Jeong Joon and Florence, Peter and Straub, Julian and Newcombe, Richard and Lovegrove, Steven},
  title     = {{DeepSDF}: Learning Continuous Signed Distance Functions for Shape Representation},
  booktitle = {Proceedings of the IEEE/CVF Conference on Computer Vision and Pattern Recognition (CVPR)},
  year      = {2019}
}

@article{zhang2023vecset,
  author    = {Zhang, Biao and Tang, Jiapeng and Nie{\ss}ner, Matthias and Wonka, Peter},
  title     = {{3DShape2VecSet}: A {3D} Shape Representation for Neural Fields and Generative Diffusion Models},
  journal   = {ACM Transactions on Graphics},
  volume    = {42},
  number    = {4},
  year      = {2023},
  doi       = {10.1145/3592442}
}

@inproceedings{peng2020convonet,
  author    = {Peng, Songyou and Niemeyer, Michael and Mescheder, Lars and Pollefeys, Marc and Geiger, Andreas},
  title     = {Convolutional Occupancy Networks},
  booktitle = {European Conference on Computer Vision (ECCV)},
  series    = {Lecture Notes in Computer Science},
  volume    = {12348},
  pages     = {523--540},
  publisher = {Springer},
  year      = {2020},
  doi       = {10.1007/978-3-030-58580-8_31}
}

@inproceedings{chibane2020ifnet,
  author    = {Chibane, Julian and Alldieck, Thiemo and Pons-Moll, Gerard},
  title     = {Implicit Functions in Feature Space for {3D} Shape Reconstruction and Completion},
  booktitle = {Proceedings of the IEEE/CVF Conference on Computer Vision and Pattern Recognition (CVPR)},
  year      = {2020}
}

@article{wang2022dog,
  author    = {Wang, Peng-Shuai and Liu, Yang and Tong, Xin},
  title     = {Dual Octree Graph Networks for Learning Adaptive Volumetric Shape Representations},
  journal   = {ACM Transactions on Graphics},
  volume    = {41},
  number    = {4},
  articleno = {103},
  year      = {2022},
  doi       = {10.1145/3528223.3530087}
}

@inproceedings{jiang2020lig,
  author    = {Jiang, Chiyu ``Max'' and Sud, Avneesh and Makadia, Ameesh and Huang, Jingwei and Nie{\ss}ner, Matthias and Funkhouser, Thomas},
  title     = {Local Implicit Grid Representations for {3D} Scenes},
  booktitle = {Proceedings of the IEEE/CVF Conference on Computer Vision and Pattern Recognition (CVPR)},
  year      = {2020}
}

@inproceedings{chabra2020deepls,
  author    = {Chabra, Rohan and Lenssen, Jan E. and Ilg, Eddy and Schmidt, Tanner and Straub, Julian and Lovegrove, Steven and Newcombe, Richard},
  title     = {Deep Local Shapes: Learning Local {SDF} Priors for Detailed {3D} Reconstruction},
  booktitle = {European Conference on Computer Vision (ECCV)},
  series    = {Lecture Notes in Computer Science},
  volume    = {12374},
  pages     = {608--625},
  publisher = {Springer},
  year      = {2020},
  doi       = {10.1007/978-3-030-58526-6_36}
}

@inproceedings{erler2020points2surf,
  author    = {Erler, Philipp and Guerrero, Paul and Ohrhallinger, Stefan and Mitra, Niloy J. and Wimmer, Michael},
  title     = {{Points2Surf}: Learning Implicit Surfaces from Point Clouds},
  booktitle = {European Conference on Computer Vision (ECCV)},
  series    = {Lecture Notes in Computer Science},
  volume    = {12350},
  pages     = {108--124},
  publisher = {Springer},
  year      = {2020},
  doi       = {10.1007/978-3-030-58558-7_7}
}

@article{erler2024ppsurf,
  author    = {Erler, Philipp and Fuentes-Perez, Lizeth and Hermosilla, Pedro and Guerrero, Paul and Pajarola, Renato and Wimmer, Michael},
  title     = {{PPSurf}: Combining Patches and Point Convolutions for Detailed Surface Reconstruction},
  journal   = {Computer Graphics Forum},
  volume    = {43},
  number    = {1},
  year      = {2024},
  doi       = {10.1111/cgf.15000}
}

@inproceedings{boulch2022poco,
  author    = {Boulch, Alexandre and Marlet, Renaud},
  title     = {{POCO}: Point Convolution for Surface Reconstruction},
  booktitle = {Proceedings of the IEEE/CVF Conference on Computer Vision and Pattern Recognition (CVPR)},
  pages     = {6302--6314},
  year      = {2022}
}

@inproceedings{li2025noksr,
  author    = {Li, Zhen and Sun, Weiwei and Govindarajan, Shrisudhan and Xia, Shaobo and Rebain, Daniel and Yi, Kwang Moo and Tagliasacchi, Andrea},
  title     = {{NoKSR}: Kernel-Free Neural Surface Reconstruction via Point Cloud Serialization},
  booktitle = {International Conference on {3D} Vision (3DV)},
  year      = {2025}
}

@inproceedings{williams2022nkf,
  author    = {Williams, Francis and Gojcic, Zan and Khamis, Sameh and Zorin, Denis and Bruna, Joan and Fidler, Sanja and Litany, Or},
  title     = {Neural Fields as Learnable Kernels for {3D} Reconstruction},
  booktitle = {Proceedings of the IEEE/CVF Conference on Computer Vision and Pattern Recognition (CVPR)},
  pages     = {18500--18510},
  year      = {2022}
}

@inproceedings{huang2023nksr,
  author    = {Huang, Jiahui and Gojcic, Zan and Atzmon, Matan and Litany, Or and Fidler, Sanja and Williams, Francis},
  title     = {Neural Kernel Surface Reconstruction},
  booktitle = {Proceedings of the IEEE/CVF Conference on Computer Vision and Pattern Recognition (CVPR)},
  pages     = {4369--4379},
  year      = {2023}
}

@inproceedings{chou2022gensdf,
  author    = {Chou, Gene and Chugunov, Ilya and Heide, Felix},
  title     = {{GenSDF}: Two-Stage Learning of Generalizable Signed Distance Functions},
  booktitle = {Advances in Neural Information Processing Systems},
  volume    = {35},
  pages     = {24905--24919},
  year      = {2022}
}

@inproceedings{ma2022osp,
  author    = {Ma, Baorui and Liu, Yu-Shen and Han, Zhizhong},
  title     = {Reconstructing Surfaces for Sparse Point Clouds with On-Surface Priors},
  booktitle = {Proceedings of the IEEE/CVF Conference on Computer Vision and Pattern Recognition (CVPR)},
  pages     = {6315--6325},
  year      = {2022}
}

@inproceedings{chen2024prior,
  author    = {Chen, Chao and Liu, Yu-Shen and Han, Zhizhong},
  title     = {Inferring Neural Signed Distance Functions by Overfitting on Single Noisy Point Clouds through Finetuning Data-Driven based Priors},
  booktitle = {Advances in Neural Information Processing Systems},
  volume    = {37},
  pages     = {60872--60896},
  year      = {2024}
}

@inproceedings{culver1999accurate,
  author    = {Culver, Tim and Keyser, John and Manocha, Dinesh},
  title     = {Accurate Computation of the Medial Axis of a Polyhedron},
  booktitle = {Proceedings of the Fifth ACM Symposium on Solid Modeling and Applications},
  pages     = {179--190},
  year      = {1999}
}

@article{culver2004exact,
  author    = {Culver, Tim and Keyser, John and Manocha, Dinesh},
  title     = {Exact Computation of the Medial Axis of a Polyhedron},
  journal   = {Computer Aided Geometric Design},
  volume    = {21},
  number    = {1},
  pages     = {65--98},
  year      = {2004}
}

@inproceedings{milenkovic1993robust,
  author    = {Milenkovic, Victor},
  title     = {Robust Construction of the {V}oronoi Diagram of a Polyhedron},
  booktitle = {Proceedings of the Canadian Conference on Computational Geometry (CCCG)},
  pages     = {473--478},
  year      = {1993}
}

@article{sherbrooke1996algorithm,
  author    = {Sherbrooke, Evan C. and Patrikalakis, Nicholas M. and Brisson, Erik},
  title     = {An Algorithm for the Medial Axis Transform of {3D} Polyhedral Solids},
  journal   = {IEEE Transactions on Visualization and Computer Graphics},
  volume    = {2},
  number    = {1},
  pages     = {44--61},
  year      = {1996}
}

@article{ramanathan2010interior,
  author    = {Ramanathan, M. and Gurumoorthy, B.},
  title     = {Interior Medial Axis Transform Computation of {3D} Objects Bound by Free-Form Surfaces},
  journal   = {Computer-Aided Design},
  volume    = {42},
  number    = {12},
  pages     = {1217--1231},
  year      = {2010}
}

@article{saha2016survey,
  author    = {Saha, Punam K. and Borgefors, Gunilla and Sanniti di Baja, Gabriella},
  title     = {A Survey on Skeletonization Algorithms and Their Applications},
  journal   = {Pattern Recognition Letters},
  volume    = {76},
  pages     = {3--12},
  year      = {2016}
}

@article{palagyi1999parallel,
  author    = {Pal{\'a}gyi, K{\'a}lm{\'a}n and Kuba, Attila},
  title     = {A Parallel {3D} 12-Subiteration Thinning Algorithm},
  journal   = {Graphical Models and Image Processing},
  volume    = {61},
  number    = {4},
  pages     = {199--221},
  year      = {1999}
}

@article{pudney1998distance,
  author    = {Pudney, Chris},
  title     = {Distance-Ordered Homotopic Thinning: A Skeletonization Algorithm for {3D} Digital Images},
  journal   = {Computer Vision and Image Understanding},
  volume    = {72},
  number    = {3},
  pages     = {404--413},
  year      = {1998}
}

@article{hesselink2008euclidean,
  author    = {Hesselink, Wim H. and Roerdink, Jos B. T. M.},
  title     = {Euclidean Skeletons of Digital Image and Volume Data in Linear Time by the Integer Medial Axis Transform},
  journal   = {IEEE Transactions on Pattern Analysis and Machine Intelligence},
  volume    = {30},
  number    = {12},
  pages     = {2204--2217},
  year      = {2008}
}

@inproceedings{rumpf2002continuous,
  author    = {Rumpf, Martin and Telea, Alexandru},
  title     = {A Continuous Skeletonization Method Based on Level Sets},
  booktitle = {Eurographics / IEEE VGTC Symposium on Visualization (VisSym)},
  editor    = {Ebert, D. and Brunet, P. and Navazo, I.},
  pages     = {151--157},
  publisher = {The Eurographics Association},
  year      = {2002},
  doi       = {10.2312/VisSym/VisSym02/151-157}
}

@article{yan2018voxel,
  author    = {Yan, Yajie and Letscher, David and Ju, Tao},
  title     = {Voxel Cores: Efficient, Robust, and Provably Good Approximation of {3D} Medial Axes},
  journal   = {ACM Transactions on Graphics},
  volume    = {37},
  number    = {4},
  year      = {2018}
}

@article{ma2012medial,
  author    = {Ma, Jaehwan and Bae, Sang Won and Choi, Sunghee},
  title     = {{3D} Medial Axis Point Approximation Using Nearest Neighbors and the Normal Field},
  journal   = {The Visual Computer},
  volume    = {28},
  pages     = {7--19},
  year      = {2012}
}

@article{jalba2013surface,
  author    = {Jalba, Andrei C. and Kustra, Jacek and Telea, Alexandru C.},
  title     = {Surface and Curve Skeletonization of Large {3D} Models on the {GPU}},
  journal   = {IEEE Transactions on Pattern Analysis and Machine Intelligence},
  volume    = {35},
  number    = {6},
  pages     = {1495--1508},
  year      = {2013}
}

@misc{huang2026offsetaxis,
  author        = {Huang, Qijia and Kraemer, Pierre and Bechmann, Dominique},
  title         = {{OffsetAxis}: {UDF} Mesh Reconstruction via Offset-Volume Medial Axis Extraction},
  year          = {2026},
  eprint        = {2605.15369},
  archivePrefix = {arXiv},
  primaryClass  = {cs.GR}
}

@article{brandt1994convergence,
  author    = {Brandt, Jonathan W.},
  title     = {Convergence and Continuity Criteria for Discrete Approximations of the Continuous Planar Skeleton},
  journal   = {CVGIP: Image Understanding},
  volume    = {59},
  number    = {1},
  pages     = {116--124},
  year      = {1994}
}

@article{amenta2001power,
  author    = {Amenta, Nina and Choi, Sunghee and Kolluri, Ravi Krishna},
  title     = {The Power Crust, Unions of Balls, and the Medial Axis Transform},
  journal   = {Computational Geometry},
  volume    = {19},
  number    = {2--3},
  pages     = {127--153},
  year      = {2001}
}

@inproceedings{attali1996modeling,
  author    = {Attali, Dominique and Montanvert, Annick},
  title     = {Modeling Noise for a Better Simplification of Skeletons},
  booktitle = {Proceedings of the 3rd IEEE International Conference on Image Processing (ICIP)},
  volume    = {3},
  pages     = {13--16},
  year      = {1996}
}

@article{dey2004approximating,
  author    = {Dey, Tamal K. and Zhao, Wulue},
  title     = {Approximating the Medial Axis from the {V}oronoi Diagram with a Convergence Guarantee},
  journal   = {Algorithmica},
  volume    = {38},
  pages     = {179--200},
  year      = {2004}
}

@inproceedings{giesen2006medial,
  author    = {Giesen, Joachim and Ramos, Edgar A. and Sadri, Bardia},
  title     = {Medial Axis Approximation and Unstable Flow Complex},
  booktitle = {Proceedings of the 22nd Annual Symposium on Computational Geometry (SoCG)},
  pages     = {327--336},
  year      = {2006}
}

@article{chazal2005lambda,
  author    = {Chazal, Fr{\'e}d{\'e}ric and Lieutier, Andr{\'e}},
  title     = {The ``$\lambda$-Medial Axis''},
  journal   = {Graphical Models},
  volume    = {67},
  number    = {4},
  pages     = {304--331},
  year      = {2005}
}

@article{chazal2008smooth,
  author    = {Chazal, Fr{\'e}d{\'e}ric and Lieutier, Andr{\'e}},
  title     = {Smooth Manifold Reconstruction from Noisy and Non-Uniform Approximation with Guarantees},
  journal   = {Computational Geometry},
  volume    = {40},
  number    = {2},
  pages     = {156--170},
  year      = {2008}
}

@inproceedings{miklos2010discrete,
  author    = {Miklos, Balint and Giesen, Joachim and Pauly, Mark},
  title     = {Discrete Scale Axis Representations for {3D} Geometry},
  booktitle = {ACM SIGGRAPH 2010 Papers},
  pages     = {1--10},
  year      = {2010}
}

@article{li2015qmat,
  author    = {Li, Pan and Wang, Bin and Sun, Feng and Guo, Xiaohu and Zhang, Caiming and Wang, Wenping},
  title     = {{Q-MAT}: Computing Medial Axis Transform by Quadratic Error Minimization},
  journal   = {ACM Transactions on Graphics},
  volume    = {35},
  number    = {1},
  pages     = {1--16},
  year      = {2015}
}

@article{pan2019qmatplus,
  author    = {Pan, Yiling and Wang, Bin and Guo, Xiaohu and Zeng, Hua and Ma, Yuexin and Wang, Wenping},
  title     = {{Q-MAT+}: An Error-Controllable and Feature-Sensitive Simplification Algorithm for Medial Axis Transform},
  journal   = {Computer Aided Geometric Design},
  volume    = {71},
  pages     = {16--29},
  year      = {2019}
}

@article{dou2022coverage,
  author    = {Dou, Zhiyang and Lin, Cheng and Xu, Rui and Yang, Lei and Xin, Shiqing and Komura, Taku and Wang, Wenping},
  title     = {Coverage Axis: Inner Point Selection for {3D} Shape Skeletonization},
  journal   = {Computer Graphics Forum},
  volume    = {41},
  pages     = {419--432},
  year      = {2022}
}

@article{wang2024coverage,
  author    = {Wang, Zimeng and Dou, Zhiyang and Xu, Rui and Lin, Cheng and Liu, Yuan and Long, Xiaoxiao and Xin, Shiqing and Komura, Taku and Yuan, Xiaoming and Wang, Wenping},
  title     = {Coverage Axis++: Efficient Inner Point Selection for {3D} Shape Skeletonization},
  journal   = {Computer Graphics Forum},
  volume    = {43},
  pages     = {e15143},
  year      = {2024}
}

@article{wang2022matfp,
  author    = {Wang, Ningna and Wang, Bin and Wang, Wenping and Guo, Xiaohu},
  title     = {Computing Medial Axis Transform with Feature Preservation via Restricted Power Diagram},
  journal   = {ACM Transactions on Graphics},
  volume    = {41},
  number    = {6},
  pages     = {1--18},
  year      = {2022}
}

@article{wang2024mattopo,
  author    = {Wang, Ningna and Huang, Hui and Song, Shibo and Wang, Bin and Wang, Wenping and Guo, Xiaohu},
  title     = {{MATTopo}: Topology-preserving Medial Axis Transform with Restricted Power Diagram},
  journal   = {ACM Transactions on Graphics},
  volume    = {43},
  number    = {6},
  articleno = {218},
  year      = {2024},
  doi       = {10.1145/3687763}
}

@inproceedings{wang2025matstruct,
  author    = {Wang, Ningna and Xu, Rui and Yin, Yibo and Zhong, Zichun and Komura, Taku and Wang, Wenping and Guo, Xiaohu},
  title     = {{MATStruct}: High-Quality Medial Mesh Computation via Structure-aware Variational Optimization},
  booktitle = {SIGGRAPH Asia 2025 Conference Papers},
  publisher = {ACM},
  year      = {2025},
  doi       = {10.1145/3757377.3763840}
}

@article{guo2024msd,
  author    = {Guo, Minghao and Wang, Bohan and Matusik, Wojciech},
  title     = {Medial Skeletal Diagram: A Generalized Medial Axis Approach for Compact {3D} Shape Representation},
  journal   = {ACM Transactions on Graphics},
  volume    = {43},
  number    = {6},
  articleno = {217},
  year      = {2024},
  doi       = {10.1145/3687964}
}

@article{tagliasacchi2009rosa,
  author    = {Tagliasacchi, Andrea and Zhang, Hao and Cohen-Or, Daniel},
  title     = {Curve Skeleton Extraction from Incomplete Point Cloud},
  journal   = {ACM Transactions on Graphics},
  volume    = {28},
  number    = {3},
  year      = {2009},
  doi       = {10.1145/1531326.1531377}
}

@article{tagliasacchi2012mcf,
  author    = {Tagliasacchi, Andrea and Alhashim, Ibraheem and Olson, Matt and Zhang, Hao},
  title     = {Mean Curvature Skeletons},
  journal   = {Computer Graphics Forum},
  volume    = {31},
  number    = {5},
  pages     = {1735--1744},
  year      = {2012},
  doi       = {10.1111/j.1467-8659.2012.03178.x}
}

@article{tagliasacchi2016star,
  author    = {Tagliasacchi, Andrea and Delame, Thomas and Spagnuolo, Michela and Amenta, Nina and Telea, Alexandru},
  title     = {{3D} Skeletons: A State-of-the-Art Report},
  journal   = {Computer Graphics Forum},
  volume    = {35},
  number    = {2},
  pages     = {573--597},
  year      = {2016},
  doi       = {10.1111/cgf.12865}
}

@article{huang2013l1,
  author    = {Huang, Hui and Wu, Shihao and Cohen-Or, Daniel and Gong, Minglun and Zhang, Hao and Li, Guiqing and Chen, Baoquan},
  title     = {{$L_1$}-Medial Skeleton of Point Cloud},
  journal   = {ACM Transactions on Graphics},
  volume    = {32},
  number    = {4},
  articleno = {65},
  year      = {2013},
  doi       = {10.1145/2461912.2461913}
}

@article{wu2015dpc,
  author    = {Wu, Shihao and Huang, Hui and Gong, Minglun and Zwicker, Matthias and Cohen-Or, Daniel},
  title     = {Deep Points Consolidation},
  journal   = {ACM Transactions on Graphics},
  volume    = {34},
  number    = {6},
  articleno = {176},
  year      = {2015},
  doi       = {10.1145/2816795.2818073}
}

@article{yang2020p2mat,
  author    = {Yang, Baorong and Yao, Junfeng and Wang, Bin and Hu, Jianwei and Pan, Yiling and Pan, Tianxiang and Wang, Wenping and Guo, Xiaohu},
  title     = {{P2MAT-NET}: Learning Medial Axis Transform from Sparse Point Clouds},
  journal   = {Computer Aided Geometric Design},
  volume    = {80},
  pages     = {101874},
  year      = {2020}
}

@inproceedings{lin2021point2skeleton,
  author    = {Lin, Cheng and Li, Changjian and Liu, Yuan and Chen, Nenglun and Choi, Yi-King and Wang, Wenping},
  title     = {{Point2Skeleton}: Learning Skeletal Representations from Point Clouds},
  booktitle = {Proceedings of the IEEE/CVF Conference on Computer Vision and Pattern Recognition (CVPR)},
  pages     = {4277--4286},
  year      = {2021}
}

@article{ge2023point2mm,
  author    = {Ge, Mengyuan and Yao, Junfeng and Yang, Baorong and Wang, Ningna and Chen, Zhonggui and Guo, Xiaohu},
  title     = {{Point2MM}: Learning Medial Mesh from Point Clouds},
  journal   = {Computers \& Graphics},
  volume    = {115},
  pages     = {511--521},
  year      = {2023}
}

@article{pepe2025dmv,
  author    = {Pepe, Antonio and Schussnig, Richard and Li, Jianning and Gsaxner, Christina and Schmalstieg, Dieter and Egger, Jan},
  title     = {Deep Medial Voxels: Learned Medial Axis Approximations for Anatomical Shape Modeling},
  journal   = {IEEE Transactions on Medical Imaging},
  volume    = {44},
  number    = {12},
  pages     = {4775--4786},
  year      = {2025},
  doi       = {10.1109/TMI.2025.3582587}
}

@inproceedings{dobbs2023smarttree,
  author    = {Dobbs, Harry and Batchelor, Oliver and Green, Richard and Atlas, James},
  title     = {{Smart-Tree}: Neural Medial Axis Approximation of Point Clouds for {3D} Tree Skeletonization},
  booktitle = {Pattern Recognition and Image Analysis (IbPRIA)},
  series    = {Lecture Notes in Computer Science},
  volume    = {14062},
  pages     = {351--362},
  publisher = {Springer},
  year      = {2023},
  doi       = {10.1007/978-3-031-36616-1_28}
}

@inproceedings{tang2019skeleton,
  author    = {Tang, Jiapeng and Han, Xiaoguang and Pan, Junyi and Jia, Kui and Tong, Xin},
  title     = {A Skeleton-Bridged Deep Learning Approach for Generating Meshes of Complex Topologies From Single {RGB} Images},
  booktitle = {Proceedings of the IEEE/CVF Conference on Computer Vision and Pattern Recognition (CVPR)},
  pages     = {4541--4550},
  year      = {2019}
}

@inproceedings{petrov2024gem3d,
  author    = {Petrov, Dmitry and Goyal, Pradyumn and Thamizharasan, Vikas and Kim, Vladimir G. and Gadelha, Matheus and Averkiou, Melinos and Chaudhuri, Siddhartha and Kalogerakis, Evangelos},
  title     = {{GEM3D}: {GE}nerative Medial Abstractions for {3D} Shape Synthesis},
  booktitle = {ACM SIGGRAPH 2024 Conference Papers},
  publisher = {ACM},
  year      = {2024},
  doi       = {10.1145/3641519.3657415}
}

@article{clemot2023neural,
  author    = {Cl{\'e}mot, Matt{\'e}o and Digne, Julie},
  title     = {Neural Skeleton: Implicit Neural Representation Away from the Surface},
  journal   = {Computers \& Graphics},
  volume    = {114},
  pages     = {368--378},
  year      = {2023}
}

@misc{rebain2021deep,
  author        = {Rebain, Daniel and Li, Ke and Sitzmann, Vincent and Yazdani, Soroosh and Yi, Kwang Moo and Tagliasacchi, Andrea},
  title         = {Deep Medial Fields},
  year          = {2021},
  eprint        = {2106.03804},
  archivePrefix = {arXiv}
}

@article{kong2026qmdf,
  author    = {Kong, Jiayi and Zong, Chen and Luo, Jun and Xin, Shiqing and Hou, Fei and Jiang, Hanqing and Qian, Chen and He, Ying},
  title     = {Quasi-Medial Distance Field ({Q-MDF}): A Robust Method for Approximating and Discretizing Neural Medial Axes},
  journal   = {ACM Transactions on Graphics},
  volume    = {45},
  number    = {3},
  year      = {2026},
  doi       = {10.1145/3795772}
}

@article{weidemaier2026maaware,
  author    = {Weidemaier, Samuel and Norden-Smoch, Christoph and Rumpf, Martin},
  title     = {Medial Axis Aware Learning of Signed Distance Functions},
  journal   = {Computer Graphics Forum},
  pages     = {e70517},
  year      = {2026},
  doi       = {10.1111/cgf.70517}
}

@incollection{blum1967transformation,
  author    = {Blum, Harry},
  title     = {A Transformation for Extracting New Descriptors of Shape},
  booktitle = {Models for the Perception of Speech and Visual Form},
  editor    = {Wathen-Dunn, Weiant},
  publisher = {MIT Press},
  address   = {Cambridge, MA},
  pages     = {362--380},
  year      = {1967}
}

@article{pizer2003mreps,
  author  = {Pizer, Stephen M. and Fletcher, P. Thomas and Joshi, Sarang and Thall, Andrew and Chen, James Z. and Fridman, Yonatan and Fritsch, Daniel S. and Gash, Graham and Glotzer, John M. and Jiroutek, Michael R. and Lu, Conglin and Muller, Keith E. and Tracton, Gregg and Yushkevich, Paul and Chaney, Edward L.},
  title   = {Deformable {M}-Reps for {3D} Medical Image Segmentation},
  journal = {International Journal of Computer Vision},
  volume  = {55},
  number  = {2--3},
  pages   = {85--106},
  year    = {2003},
  doi     = {10.1023/A:1026313132218}
}

@article{damon2003skeletal,
  author  = {Damon, James},
  title   = {Smoothness and Geometry of Boundaries Associated to Skeletal Structures {I}: Sufficient Conditions for Smoothness},
  journal = {Annales de l'Institut Fourier},
  volume  = {53},
  number  = {6},
  pages   = {1941--1985},
  year    = {2003},
  doi     = {10.5802/aif.1997}
}

@inproceedings{qi2017pointnet,
  author    = {Qi, Charles R. and Su, Hao and Mo, Kaichun and Guibas, Leonidas J.},
  title     = {{PointNet}: Deep Learning on Point Sets for {3D} Classification and Segmentation},
  booktitle = {Proceedings of the IEEE Conference on Computer Vision and Pattern Recognition (CVPR)},
  pages     = {652--660},
  year      = {2017}
}

@inproceedings{guerrero2018pcpnet,
  title={Pcpnet learning local shape properties from raw point clouds},
  author={Guerrero, Paul and Kleiman, Yanir and Ovsjanikov, Maks and Mitra, Niloy J},
  booktitle={Computer graphics forum},
  volume={37},
  number={2},
  pages={75--85},
  year={2018},
  organization={Wiley Online Library}
}

@inproceedings{koch2019abc,
  title={Abc: A big cad model dataset for geometric deep learning},
  author={Koch, Sebastian and Matveev, Albert and Jiang, Zhongshi and Williams, Francis and Artemov, Alexey and Burnaev, Evgeny and Alexa, Marc and Zorin, Denis and Panozzo, Daniele},
  booktitle={2019 IEEE/CVF conference on computer vision and pattern Recognition (CVPR)},
  pages={9593--9603},
  year={2019},
  organization={Ieee}
}

@inproceedings{ling2022vectoradam,
  author    = {Ling, Selena and Sharp, Nicholas and Jacobson, Alec},
  title     = {{VectorAdam} for Rotation Equivariant Geometry Optimization},
  booktitle = {Advances in Neural Information Processing Systems (NeurIPS)},
  volume    = {35},
  year      = {2022}
}

@article{hou2023dcudf,
  author    = {Hou, Fei and Chen, Xuhui and Wang, Wencheng and Qin, Hong and He, Ying},
  title     = {Robust Zero Level-Set Extraction from Unsigned Distance Fields Based on Double Covering},
  journal   = {ACM Trans. Graph.},
  volume    = {42},
  number    = {6},
  year      = {2023},
  note      = {SIGGRAPH Asia 2023}
}
